\documentclass[runningheads]{llncs}

\usepackage{eccv}

\usepackage{eccvabbrv}
\usepackage{subcaption}
\usepackage{bm}
\usepackage{graphicx}
\usepackage{booktabs}
\usepackage{multirow}
\usepackage[accsupp]{axessibility}  % Improves PDF readability for those with disabilities.

\usepackage{hyperref}
\usepackage{orcidlink}

\begin{document}

% --------------------------------------------------------------- 
\title{RAE-NWM: Navigation World Model in Dense Visual Representation Space}

\titlerunning{RAE-NWM}

\author{
Mingkun Zhang\inst{1} \and
Wangtian Shen\inst{1} \and
Fan Zhang\inst{2} \and
Haijian Qin\inst{3} \and
Zihao Pei\inst{1} \and
Ziyang Meng\inst{1}\thanks{Corresponding author}
}

\authorrunning{M. Zhang et al.}

\institute{
Department of Precision Instrument, Tsinghua University, Beijing, China\\
\email{zhangmk24@mails.tsinghua.edu.cn, ziyangmeng@tsinghua.edu.cn}
\and
University of Rochester, Rochester, NY, USA
\and
Beijing Information Science and Technology University, Beijing, China
}

\maketitle

\begin{abstract}
Visual navigation requires agents to reach goals in complex environments through perception and planning. World models address this task by simulating action-conditioned state transitions to predict future observations. Current navigation world models typically learn state evolution under actions within the compressed latent space of a Variational Autoencoder, where spatial compression often discards fine-grained structural information and makes precise action-conditioned prediction more difficult. To better understand the propagation characteristics of different representations, we conduct a linear dynamics probe and observe that dense DINOv2 features exhibit stronger linear predictability for action-conditioned transitions. Motivated by this observation, we propose the Representation Autoencoder-based Navigation World Model (RAE-NWM), which generatively models navigation dynamics in a dense visual representation space. We employ a Conditional Diffusion Transformer with a Decoupled Diffusion Transformer head (CDiT-DH) to model continuous transitions, and introduce a separate time-driven gating module for dynamics conditioning to regulate action injection strength during generation. Extensive evaluations show that modeling sequential rollouts in this space improves structural stability and action accuracy, benefiting downstream planning and navigation. Code is available at \url{https://github.com/20robo/raenwm}.
\end{abstract}

\section{Introduction}
\label{sec:intro}

\begin{figure}[t]
  \centering
  \includegraphics[width=0.9\linewidth]{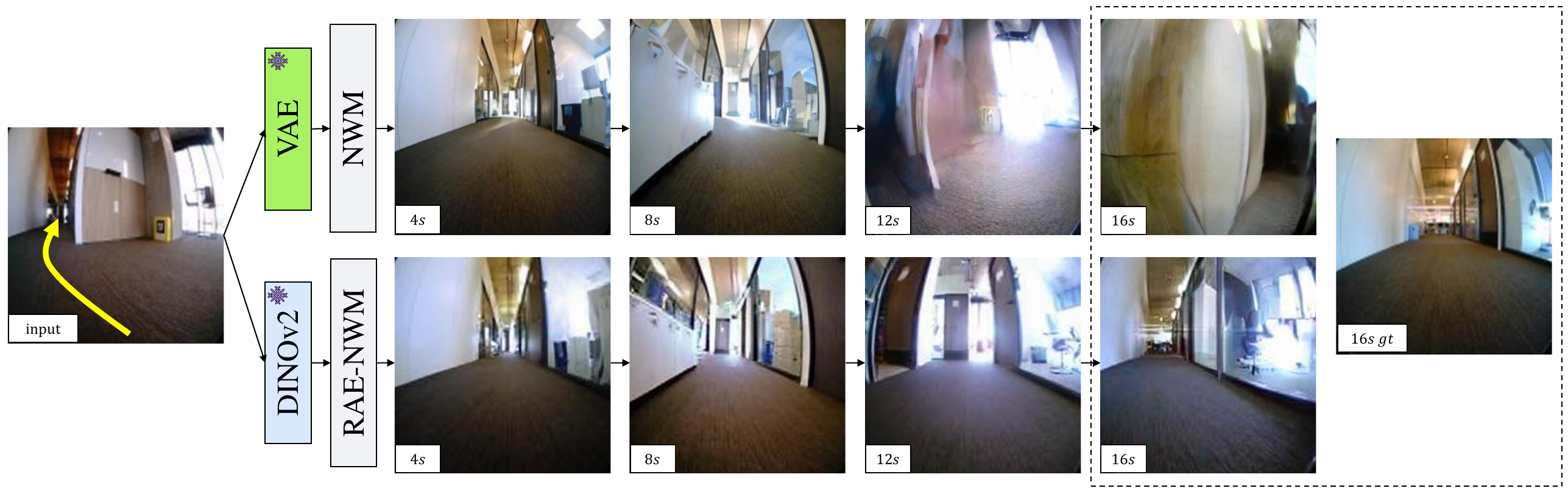}
  \caption{
  Long-horizon generation comparison.
  We compare sequential rollouts from the baseline VAE-based NWM versus our DINO-based RAE-NWM. The VAE-based model exhibits severe structural degradation at later horizons ($k=12s, 16s$), whereas RAE-NWM maintains strong structural integrity throughout the sequence.
  }
  \label{fig:intro_compare}
\end{figure}

Visual navigation is a fundamental problem in autonomous robotics, requiring agents to perceive their surroundings and reach target goals safely in complex environments~\cite{zhu2017target,duan2022survey}. End-to-end learning-based solution~\cite{sridhar2024nomad,shah2023gnm} has emerged as an effective approach for this task, but it often struggles with uninterpretable intermediate decision processes and is difficult to incorporate additional constraints after training. Navigation World Models (NWM)~\cite{bar2025_nwm}, on the other hand, offer an explicit solution with a flexible structure to address these issues by evaluating candidate trajectories through environment simulation. Using generative models conditioned on historical states and physical actions, NWM predicts future observations to assess navigation safety and goal progress. In the context of visual navigation, since these predictions directly inform how to plan, the world model must do more than simply synthesize high-fidelity images. It is crucial that the model maintains consistent spatial geometric stability and precise action controllability over extended horizons; otherwise, the reliability of the resulting decisions drops significantly. However, most existing methods~\cite{zhang2025aerial,shen2026efficient} perform these predictive rollouts within the latent space of a Variational Autoencoder (VAE)~\cite{kingma2013auto}. While such a spatial compression represents an inherent mechanism of this encoding process, it typically limits the preservation of critical geometric information from the original observations~\cite{rombach2022high,zheng2025rae}. Consequently, during long-horizon future predictions, this lack of structural consistency often causes VAE-based methods to experience significant structural collapse and kinematic deviation, limiting their reliability for downstream path planning (Figure~\ref{fig:intro_compare}).

To address the spatial degradation introduced by VAE-based latent spaces, it is important to consider representation spaces that preserve geometric structure. DINOv2~\cite{oquab2023_dinov2} contains rich spatial semantics and geometric structures, as evidenced by its strong performance in various recent vision tasks~\cite{wang2025vggt,lin2025depth}. Some recent studies~\cite{zhou2024dino,zhang2025efficient,assran2025v} attempt to use self-supervised or representative visual features for environment simulation, showing the promise of feature-space world modeling. However, existing feature-space predictive models are mainly explored through deterministic feature prediction, often in manipulation or general video prediction/planning settings, rather than generative, action-conditioned rollouts for first-person visual navigation. Continuous generative models offer a suitable framework for modeling these smooth temporal transitions. Given the inherent cost of training such heavily parameterized networks, it is prudent to first verify how well the semantic representation space supports action-conditioned dynamics. Therefore, we introduce a Linear Dynamics Probe to investigate this property. Our analysis shows that DINOv2 features exhibit strong linear predictability for action-conditioned state transitions. Inspired by this observation, we build our generative transition model entirely in this dense visual representation space.

In particular, we propose the Representation Autoencoder-based Navigation World Model (RAE-NWM) in this paper. Following the RAE~\cite{zheng2025rae} approach, our model extracts uncompressed visual tokens using a frozen DINOv2 encoder and reconstructs the final observations using a frozen pretrained RAE decoder. The decoder is used only as a fixed reconstruction component, and only the transition model between the frozen encoder and decoder is trained. Specifically, we train a Conditional Diffusion Transformer with a Decoupled Diffusion Transformer head (CDiT-DH) to serve as the generative backbone~\cite{wang2025ddt,peebles2023scalable}. Motivated by the coarse-to-fine nature of diffusion-style generative processes~\cite{balaji2022ediff,hertz2022prompt}, we introduce a dynamics conditioning module equipped with a time-driven dynamic gating mechanism. Instead of standard additive conditioning, this module adaptively adjusts the strength of kinematic conditioning throughout the probability flow. This mechanism balances high-fidelity visual generation and precise action controllability, preserving global geometric topology while refining local structural details.

Our contributions can be summarized as follows.
First, we study generative navigation world modeling in dense visual representation spaces instead of compressed VAE-based latent spaces. This formulation preserves richer spatial structure and provides a suitable representation space for modeling action-conditioned first-person navigation dynamics.
Second, we develop a generative transition architecture for navigation world models built on CDiT-DH and an adaptive gating mechanism. This design enables stable modeling of high-dimensional visual representations while maintaining global geometric consistency and local structural details.
Third, extensive experiments verify that our approach improves long-horizon rollout stability and achieves stronger performance in both open-loop trajectory evaluation and downstream navigation planning tasks.

\section{Related Work}

\paragraph{Visual Representation Space.}
Traditional generative paradigms typically construct their representation space using compression models. Standard approaches, from the foundational Variational Autoencoder (VAE)~\cite{kingma2013auto,rombach2022high} to methods incorporating semantic alignments or masked reconstruction objectives~\cite{yao2025reconstruction,chen2025masked,chen2025dc,yang2025latent}, introduce a spatial compression bottleneck. Relying on these low-dimensional latents limits low-level geometric fidelity, even with semantic enhancements. To address this limitation, Representation Autoencoders (RAE)~\cite{zheng2025rae} reconstruct images directly from uncompressed visual foundation models~\cite{oquab2023_dinov2,tschannen2025siglip,he2022masked}. Although such representations are often considered to primarily encode high-level semantic information rather than low-level visual details~\cite{tang2025unilip,yu2024image}, RAE shows that they can nevertheless support high-fidelity spatial reconstruction when paired with a vision transformer (ViT)~\cite{dosovitskiy2020image} decoder, without relying on compression.

\paragraph{World Models for Embodied Intelligence.}
Unlike end-to-end visual navigation policies~\cite{sridhar2024nomad,shah2023gnm,shen2025effonav,shah2023vint,shah2022viking,qin2025esem}, world models~\cite{ha2018world, li2025comprehensive} explicitly simulate environmental dynamics for decision-making. Recent Navigation World Models~\cite{bar2025_nwm,shen2026efficient} condition future predictions on physical actions. However, these frameworks often rely on VAE-based latent spaces, which can degrade spatial structure. To mitigate this issue, some methods~\cite{zhang2025aerial,yu2024viewcrafter} explicitly project past frames into future views, but this increases computational complexity. Other works explore hybrid architectures that combine autoregressive generation with diffusion chunking~\cite{shang2025longscape} to extend rollout horizons, or study representation-space world models based on self-supervised or representative visual features~\cite{zhou2024dino,zhang2025efficient,assran2025v}. DINO-WM~\cite{zhou2024dino} learns dynamics in continuous DINOv2 feature space through deterministic feature prediction, while V-JEPA 2~\cite{assran2025v} studies self-supervised video prediction and planning. Cloning Deterministic Worlds~\cite{xia2026cloning} further discusses limitations of pixel-reconstructing VAE representations for navigation world modeling. These works demonstrate the potential of representation-level simulation, but they mostly rely on deterministic prediction or settings such as manipulation and general video prediction, rather than generative first-person navigation rollouts. Since navigation dynamics follow smooth action-induced geometric changes, RAE-NWM models them with continuous-time generative dynamics in dense representation space for stable long-horizon planning.

\section{Task Definition and Representation Analysis}
\label{sec:task_and_probe}

\subsection{Task Definition}
\label{sec:task_def}
A world model couples an encoder, which maps complex raw observations into a tractable state space, with a predictor that forecasts future states based on agent actions. In planar visual navigation tasks, an agent receives a first-person RGB observation $\mathbf{o}_i \in \mathbb{R}^{H \times W \times 3}$ at step $i$. We map a sequence of $m$ context frames $\mathbf{o}_{\mathrm{cond},i} = \mathbf{o}_{i-m+1:i}$ through an encoder to obtain the state representation $\mathbf{z}_{\mathrm{cond},i} = \mathrm{Enc}(\mathbf{o}_{\mathrm{cond},i})$, capturing the current environmental context.
Following NWM~\cite{bar2025_nwm}, we introduce a relative temporal shift $k$ to specify the prediction horizon. We then represent the agent motion over this interval using a planar motion condition $\bm{a}_{i \rightarrow i+k} = [u_x, u_y, \omega]_{i \rightarrow i+k}$, where $u_x, u_y$ and $\omega$ denote the planar translation and yaw rotation respectively. Together, $k$ and $\bm{a}_{i \rightarrow i+k}$ describe the kinematic condition of the agent during this period. The objective of the world model for visual navigation tasks is to learn a transition dynamics function within this representation space:
\begin{equation}
\hat{\mathbf{z}}_{i+k} = \mathrm{Pred}_{\theta}(\mathbf{z}_{\mathrm{cond},i}, \bm{a}_{i \rightarrow i+k}, k)
\label{eq:task_def}
\end{equation}

\begin{figure}[t]
  \centering
  \includegraphics[width=0.4\linewidth]{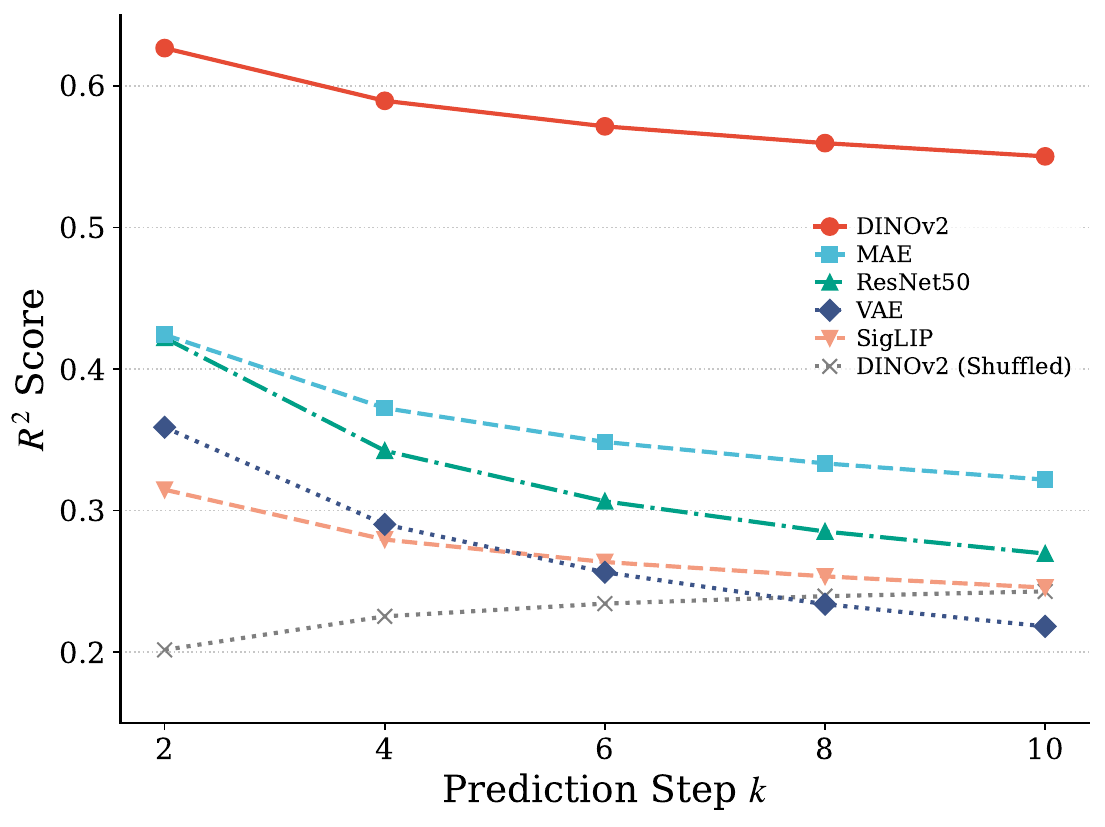}
  \caption{
  Linear predictability of action-conditioned state transitions across various visual representation spaces, measured by the global $R^2$ score over the prediction horizon step $k$.
  }
  \label{fig:probe_r2}
\end{figure}

\subsection{Representation Analysis}
\label{sec:probe}

To effectively learn the transition function defined in Equation~\ref{eq:task_def}, it is critical to determine an appropriate state representation space to encode raw observations. This choice determines whether the subsequent predictor can readily extract action-conditioned dynamics from the representation, or instead has to model complex visual variations that are unrelated to the underlying motion. Inspired by linear probing techniques~\cite{alain2016understanding}, we introduce a linear dynamics probe to evaluate the kinematic predictability within various representation spaces. To evaluate such predictability without involving generative modeling, we freeze the pre-trained encoder and only train a linear probe in its representations. We extract the spatial token representation $\mathbf{z}_{i}$ from the current observation and use the future representation $\mathbf{z}_{i+k}$ as the prediction target. The probe predicts the future state conditioned on the intermediate motion $\bm{a}_{i \rightarrow i+k}$ through a linear formulation:
\begin{equation}
    \hat{\mathbf{z}}_{i+k} = \mathbf{z}_i + \bm{a}(\mathbf{z}_i) + \mathbf{B}(\bm{a}_{i \rightarrow i+k})
    \label{eq:linear_probe_formulation}
\end{equation}
where $\bm{a}$ and $\mathbf{B}$ denote learnable linear transformations. We restrict the probe to this simple setup to ensure the results directly reveal whether the underlying feature space natively supports action-conditioned dynamics.

The linear probes are trained and evaluated on the navigation datasets detailed in Section~\ref{subsec:experimental setup}. To quantify this linear predictability, we report the global $R^2$ score~\cite{hastie2009elements}, which measures how well future states can be explained by a linear function of the current state and action. As indicated by Figure~\ref{fig:probe_r2}, the uncompressed token space of DINOv2 maintains consistently high predictability across the entire prediction horizon. In contrast, VAE-based compressed representations yield substantially lower scores. Other common visual encoders, such as those optimized for pixel-level masked reconstruction (e.g., MAE~\cite{he2022masked}) or global semantic alignment (e.g., SigLIP~\cite{tschannen2025siglip} and ResNet50~\cite{he2016deep}), also perform poorly. Furthermore, to study whether the high scores of DINO tokens arise mainly from its representation capability of global semantic information, we introduce a spatially shuffled DINOv2 baseline, where the spatial order of the encoded tokens is randomly permuted before being fed to the probe. This operation disrupts the spatial structure of the representation, and the resulting performance drops substantially compared to the original DINO features. This observation suggests that the performance gap comes from the spatial structure of DINOv2 features, rather than static global semantics alone. Therefore, we build the world model directly in this dense representation space.

\begin{figure}[t]
  \centering
  \includegraphics[width=0.9\linewidth]{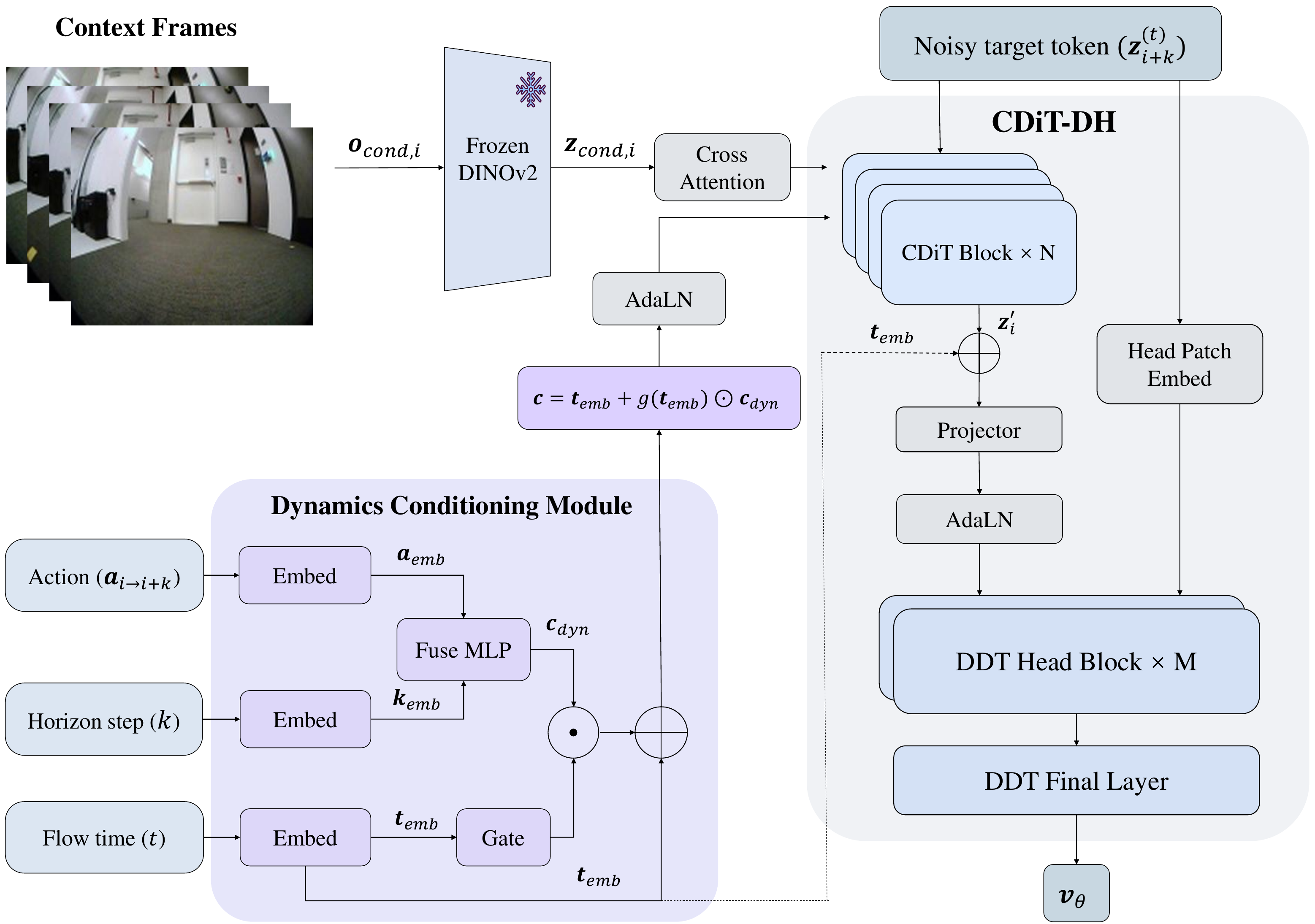}
    \caption{Architecture of the proposed RAE-NWM. Our model encodes context frames into a sequence of tokens $\mathbf{z}_{\mathrm{cond},i}$ via a frozen DINOv2. The dynamics conditioning module then integrates the agent motion $\bm{a}_{i\rightarrow i+k}$ and prediction horizon $k$ with the flow time $t$ through a time-driven gating mechanism. Finally, the CDiT-DH backbone predicts the flow velocity $\mathbf{v}_\theta$.}
  \label{fig:model}
\end{figure}

\section{RAE-NWM}
\label{sec:rae-nwm}

We formulate the proposed Representation Autoencoder-based Navigation World Model (RAE-NWM) to parameterize the transition dynamics function defined in Section~\ref{sec:task_and_probe}. This section details the network architecture and the training pipeline. We first specify the visual state representation extraction (Sec.~\ref{sec:state_rep}) and introduce the generative backbone CDiT-DH (Sec.~\ref{sec:cditdh}). We then describe the time-driven dynamics conditioning module designed for adaptive kinematic modulation (Sec.~\ref{sec:cond}), and finally describe the training objective and sequential rollout procedure (Sec.~\ref{sec:training_inference}).

\subsection{State Representation}
\label{sec:state_rep}

Given a raw visual observation $\mathbf{o}_i$, we construct its state representation strictly within the dense representation space. We employ a frozen DINOv2 encoder and discard the \texttt{[CLS]} token to retain only the uncompressed spatial patch tokens $\mathbf{z}_i = \mathrm{Enc}(\mathbf{o}_i) \in \mathbb{R}^{L \times d}$. In our implementation, we extract $L = 16 \times 16 = 256$ patch tokens with a feature dimension of $d = 768$. To incorporate temporal context, the features from the context frames are organized into the sequence $\mathbf{z}_{\mathrm{cond},i} = \mathbf{z}_{i-m+1:i}$. This sequence serves as the spatial conditioning input for the generative backbone.

\subsection{Generative Backbone: CDiT-DH}
\label{sec:cditdh}

As illustrated in Figure~\ref{fig:model}, after extracting the context representation $\mathbf{z}_{\mathrm{cond},i}$, we employ the CDiT-DH generative backbone. Following the flow matching formulation~\cite{lipman2022flow}, the CDiT-DH takes as input a noisy target token $\mathbf{z}_{i+k}^{(t)}$. Here $t \in [0,1]$ denotes the continuous flow time, and the network predicts the corresponding velocity field $\mathbf{v}_\theta$. This prediction is conditioned on the context tokens $\mathbf{z}_{\mathrm{cond},i}$ and a unified global condition vector $\mathbf{c}$ for action and horizon-step control (detailed in Section~\ref{sec:cond}).

\paragraph{Deep Generative Backbone.}
Following the design of NWM~\cite{bar2025_nwm}, we use a deep Conditional Diffusion Transformer (CDiT) composed of $N$ stacked transformer blocks. Given the noisy target token $\mathbf{z}_{i+k}^{(t)}$, the backbone embeds it into patch tokens and adds a 2D sine-cosine positional embedding~\cite{he2022masked} to provide an explicit spatial inductive bias. Each block applies self-attention to model token-wise spatial dependencies and cross-attention to incorporate the context frames, attending to $\mathbf{z}_{\mathrm{cond},i}$ as keys and values. The global condition $\mathbf{c}$ is injected via Adaptive Layer Normalization (AdaLN)~\cite{peebles2023scalable} to modulate features throughout the transformer blocks. The generative backbone produces a deep contextual feature $\mathbf{z}'_i$:
\begin{equation}
\mathbf{z}'_i = \mathrm{CDiT}_{\theta}\!\left(\mathbf{z}_{i+k}^{(t)} \mid \mathbf{z}_{\mathrm{cond},i},\, \mathbf{c}\right).
\label{eq:cdit_backbone}
\end{equation}

\paragraph{Decoupled Diffusion Transformer (DDT) Head.}
Modeling target representations in a high-dimensional token space can be challenging for standard diffusion transformer backbones. Following the design principle of RAE~\cite{zheng2025rae}, we adopt a shallow-and-wide DDT~\cite{wang2025ddt} head to predict the final velocity field. This lightweight head allows the model to better handle high-dimensional token representations without significantly increasing computational cost. In practice, the re-embedded noisy input is modulated by spatial AdaLN under the guidance of the deep features produced by the backbone. The predicted velocity field is then computed as
\begin{equation}
\mathbf{v}_{\theta}
=
\mathrm{DDT}_{\theta}
\big(
\mathbf{z}_{i+k}^{(t)}
\mid
\mathbf{z}'_i, t
\big).
\label{eq:ddt_head}
\end{equation}

\subsection{Dynamics Conditioning Module}
\label{sec:cond}

To condition the generative backbone while preserving the temporal structure of the generative process, we introduce a dynamics conditioning module that adaptively injects the kinematic control signal. We apply Gaussian Fourier embedding~\cite{song2020score} to map the input action $\bm{a}_{i\rightarrow i+k}$, the horizon step $k$, and the continuous flow time $t$ into a high-frequency latent space. This produces the embeddings $\bm{a}_{\mathrm{emb}}$, $\mathbf{k}_{\mathrm{emb}}$, and $\mathbf{t}_{\mathrm{emb}}$. We then concatenate $\bm{a}_{\mathrm{emb}}$ and $\mathbf{k}_{\mathrm{emb}}$ and process them through a multi-layer perceptron (MLP) to extract the dynamics feature $\mathbf{c}_{\mathrm{dyn}}$, which captures the intended motion.

To better balance precise action control and fine-grained visual synthesis within the high-dimensional DINOv2 space, we replace standard additive injection with a time-driven gating function $g(\mathbf{t}_{\mathrm{emb}})$ to dynamically modulate the conditioning strength. This adaptive design aligns with the generation process. Early high-noise stages need strong kinematic priors to establish the global topology, while late low-noise stages require relaxed constraints to refine high-frequency visual details without introducing artifacts. The function $g(\cdot)$ consists of a linear layer and a SiLU activation, followed by a Sigmoid function to constrain the outputs to $(0, 1)$. The final global condition vector $\mathbf{c}$ is formulated as:
\begin{equation}
\mathbf{c} = \mathbf{t}_{\mathrm{emb}} + g(\mathbf{t}_{\mathrm{emb}}) \odot \mathbf{c}_{\mathrm{dyn}}
\label{eq:global_cond}
\end{equation}
where $\odot$ denotes element-wise multiplication. This modulated condition $\mathbf{c}$ is then injected into the network via AdaLN to guide spatial generation.

\begin{figure}[t]
  \centering
  \includegraphics[width=0.9\linewidth]{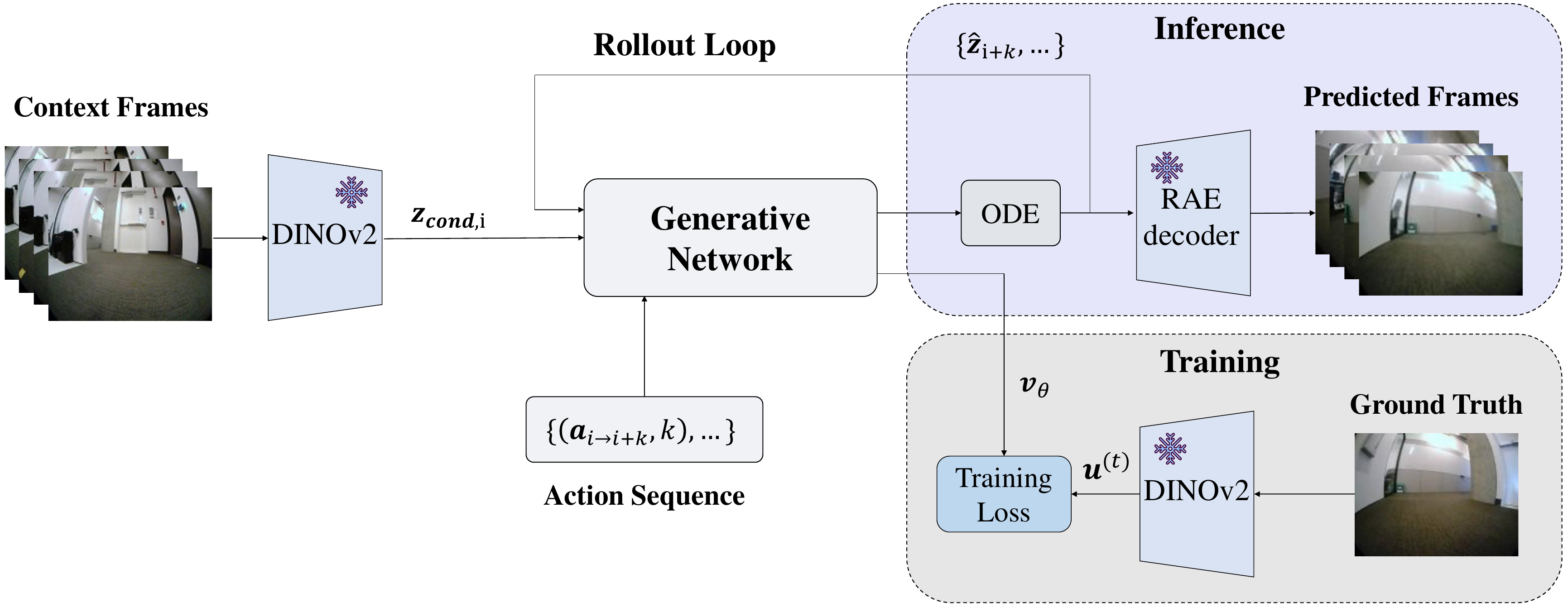}
\caption{
  Training and sequential rollout pipelines of RAE-NWM. 
  During training, the generative network is optimized via a flow matching objective to predict the velocity field $\mathbf{v}_\theta$ matching the target velocity $\mathbf{u}^{(t)}$. 
  During inference, an ordinary differential equation (ODE) solver sequentially generates future states $\{\hat{\mathbf{z}}_{i+k}, \dots\}$ along a given action sequence $\{(\bm{a}_{i\rightarrow i+k}, k), \dots\}$ within a closed representation-space rollout loop. 
  The frozen RAE decoder is applied exclusively for qualitative visualization and pixel-level metric evaluation.
}
  \label{fig:rollout}
\end{figure}

\subsection{Training Objective and Sequential Rollout}
\label{sec:training_inference}

Figure~\ref{fig:rollout} summarizes the training and sequential rollout pipelines of RAE-NWM. This section details the Flow Matching~\cite{lipman2022flow} training objective and the representation-space rollout procedure for long-horizon prediction. For clarity, we denote the target representation at a specific horizon as $\mathbf{z}^{(t)}$ and the historical context as $\mathbf{z}_{\mathrm{cond}}$, where $t$ is the flow time. During training, we construct a continuous probability path that transports the clean representation $\mathbf{z}^{(0)}$ to standard Gaussian noise $\mathbf{z}^{(1)} \sim \mathcal{N}(\mathbf{0}, \mathbf{I})$. We define a linear interpolation path for any $t \in [0, 1]$ as $\mathbf{z}^{(t)} = (1-t)\mathbf{z}^{(0)} + t\mathbf{z}^{(1)}$, which yields the target velocity field $\mathbf{u}^{(t)} = \frac{d\mathbf{z}^{(t)}}{dt} = \mathbf{z}^{(1)} - \mathbf{z}^{(0)}$. The generative network is optimized by matching the predicted velocity $\mathbf{v}_\theta$ directly to this target:
\begin{equation}
\mathcal{L} = \mathbb{E}_{\mathbf{z}^{(0)},\,\mathbf{z}^{(1)},\,t}\Big[ \big\| \mathbf{v}_\theta(\mathbf{z}^{(t)}, t, \mathbf{z}_{\mathrm{cond}}, \mathbf{c}) - \mathbf{u}^{(t)} \big\|_2^2 \Big].
\label{eq:fm_loss}
\end{equation}

During inference for the visual navigation task, long-horizon prediction is executed sequentially within the dense representation space. After encoding the initial context frames to obtain the context tokens $\mathbf{z}_{\mathrm{cond},i}$, we initialize the target token for each prediction interval with standard Gaussian noise $\mathbf{z}^{(1)}$. An ordinary differential equation (ODE) solver computes the probability flow backward from $t=1$ to $t=0$, guided by the context and the dynamics condition $\mathbf{c}$, producing the predicted clean token $\hat{\mathbf{z}}_{i+k}$. To predict further into the future along a given action sequence, this newly generated state is appended to the sliding context window to condition the subsequent rollout step. The frozen pre-trained RAE decoder~\cite{zheng2025rae} reconstructs these representations into pixel space exclusively for qualitative visualization and pixel-level metric evaluation. For trajectory scoring and planning, we operate directly in the representation space instead of decoding predictions to pixels, avoiding geometric distortions and information loss from pixel reconstruction.

\section{Experiments}
\subsection{Experimental Setup}
\label{subsec:experimental setup}

\paragraph{Datasets.}
We evaluate our approach on three complementary real-world robot navigation datasets: SACSoN/HuRoN~\cite{hirose2023sacson}, RECON~\cite{shah2021rapid}, and SCAND~\cite{karnan2022scand}. SACSoN captures dynamic indoor human-robot interactions, RECON provides unstructured open-world and off-road terrains, and SCAND contains demonstrations that follow human social norms. We train a single model on the union of their training splits and report results on held-out trajectories from each dataset. We also collect 1{,}000 trajectories in Matterport3D~\cite{chang2017matterport3d} using the Habitat simulator~\cite{szot2021habitat}. Models for the Habitat experiments are trained exclusively on this dataset and evaluated on unseen trajectories. All datasets are partitioned at the trajectory level to prevent temporal leakage.

\paragraph{Evaluation Metrics.}
To assess open-loop generation quality, we report the Fréchet Inception Distance (FID)~\cite{heusel2017gans} for frame-level realism and measure perceptual similarity using LPIPS~\cite{zhang2018unreasonable} and DreamSim~\cite{fu2023dreamsim}. We also introduce a patch-wise DINO feature distance, computed as the mean cosine distance between normalized patch tokens extracted by a frozen DINOv2, to evaluate dense semantics and geometric consistency. To measure trajectory and planning accuracy, we test the utility of predicted observations in a downstream planning task based on Cross-Entropy Method, quantifying trajectory fidelity with Absolute Trajectory Error (ATE) and Relative Pose Error (RPE)~\cite{sturm2012evaluating}. Finally, to evaluate closed-loop simulation performance in Habitat, we deploy the model as a robotic navigation system and report Success Rate (SR) and Success weighted by Path Length (SPL)~\cite{anderson2018evaluation}.

\paragraph{Baselines.}
We compare our approach against several representative baselines. NWM~\cite{bar2025_nwm} is a world-modeling framework for real-world visual navigation and serves as our primary baseline for open-loop generation and downstream planning. GNM~\cite{shah2023gnm} and NoMaD~\cite{sridhar2024nomad} are end-to-end visual navigation policies trained on multiple datasets. GNM is a goal-conditioned policy, while NoMaD introduces a diffusion policy to unify exploration and navigation. For the simulation-oriented Habitat evaluation, we compare against OmniVLA~\cite{hirose2025omnivla}, which builds upon the OpenVLA~\cite{kim2024openvla} backbone, and the One-Step World Model (One-Step WM)~\cite{shen2026efficient}, which uses a 3D U-Net dynamics model for one-step generation.

\paragraph{Implementation Details.}
Detailed network architecture hyperparameters, training configurations, and computational resources are provided in Section A of the Supplementary Material.

\subsection{Open-Loop Generation}
\label{subsec:open_loop}

\begin{table}[t]
\centering
\caption{Direct long-horizon prediction quality. Performance comparison between RAE-NWM and NWM on the SACSoN dataset at 4-second and 16-second future horizons, generated directly in a single step bypassing intermediate frames.}
\label{tab:future_prediction}
\begin{tabular}{lcccccccc}
\toprule
\multirow{2}{*}{Model} & \multicolumn{2}{c}{LPIPS $\downarrow$} & \multicolumn{2}{c}{DreamSim $\downarrow$} & \multicolumn{2}{c}{DINO Distance $\downarrow$} & \multicolumn{2}{c}{FID $\downarrow$} \\
\cmidrule(lr){2-3} \cmidrule(lr){4-5} \cmidrule(lr){6-7} \cmidrule(lr){8-9}
& 4s & 16s & 4s & 16s & 4s & 16s & 4s & 16s \\
\midrule
NWM & 0.407 & 0.470 & 0.229 & 0.281 & 0.402 & 0.460 & 26.15 & 33.06 \\
RAE-NWM & \textbf{0.303} & \textbf{0.349} & \textbf{0.145} & \textbf{0.171} & \textbf{0.327} & \textbf{0.367} & \textbf{15.09} & \textbf{15.90} \\
\bottomrule
\end{tabular}
\end{table}

\begin{figure}[t]
  \centering
  \includegraphics[width=0.9\linewidth]{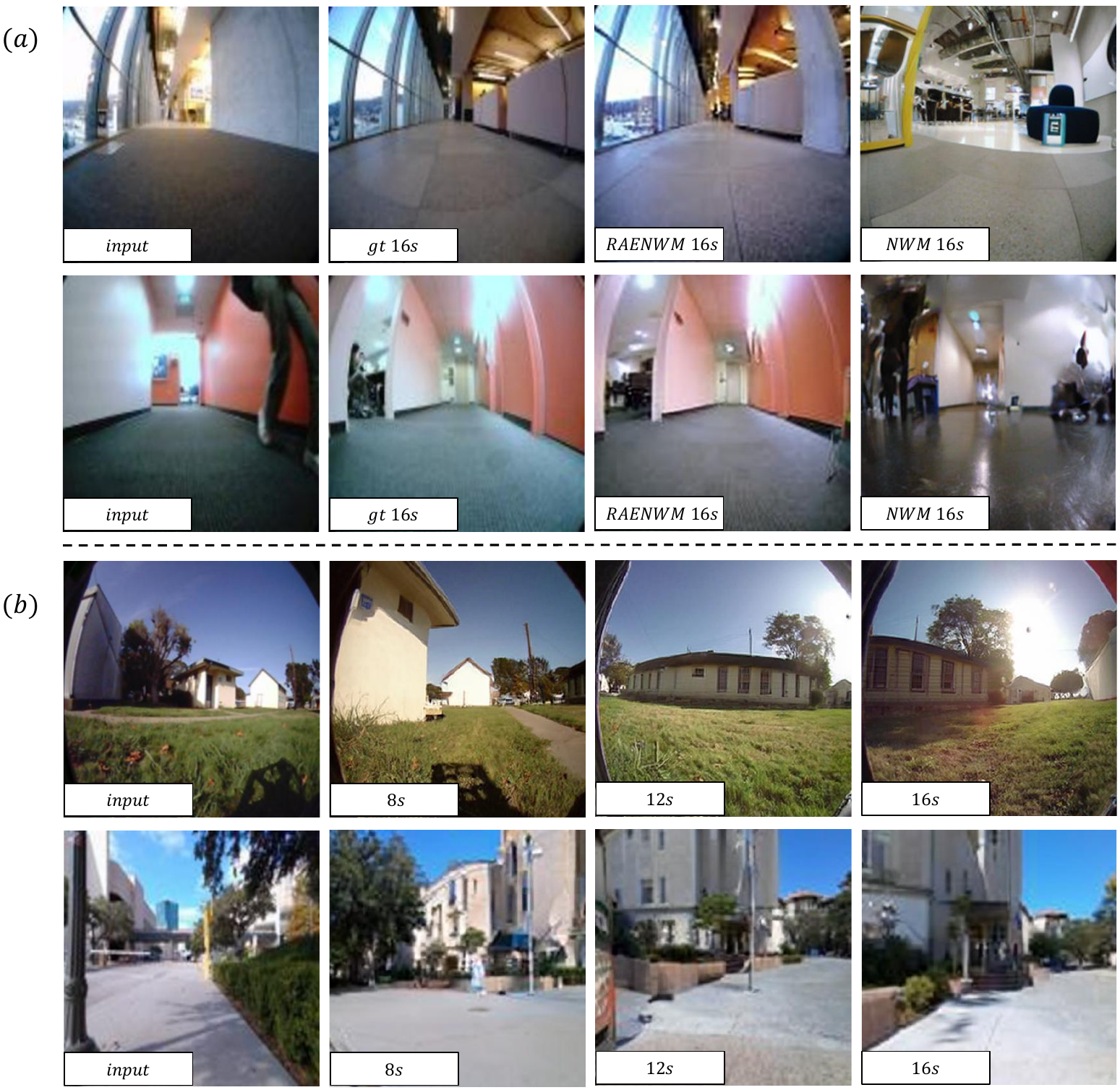}
  \caption{
  Qualitative results of long-horizon generation. 
  (a) Direct prediction at the 16-second horizon. NWM produces observations that completely deviate from the ground truth, whereas RAE-NWM maintains high geometric fidelity, demonstrating superior action-conditioned dynamics. 
  (b) Sequential rollouts of RAE-NWM on the RECON and SCAND datasets, exhibiting strong structural consistency and spatial stability over extended horizons.
  }
  \label{fig:qualitative}
\end{figure}

To identify long-horizon action controllability from the error accumulation inherent in sequential rollout, we first perform a direct prediction experiment. Instead of iteratively generating intermediate frames, we predict specific future observations directly at 4-second and 16-second horizons based on aggregated action sequences. Table~\ref{tab:future_prediction} demonstrates that NWM degrades significantly across perceptual and semantic metrics at the 16-second mark. As qualitatively shown in Figure~\ref{fig:qualitative}a, NWM produces observations that deviate substantially from the ground truth, whereas RAE-NWM preserves much higher geometric fidelity.

\begin{figure*}[t]
    \centering
    \begin{subfigure}[b]{0.24\textwidth}
        \centering
        \includegraphics[width=\textwidth]{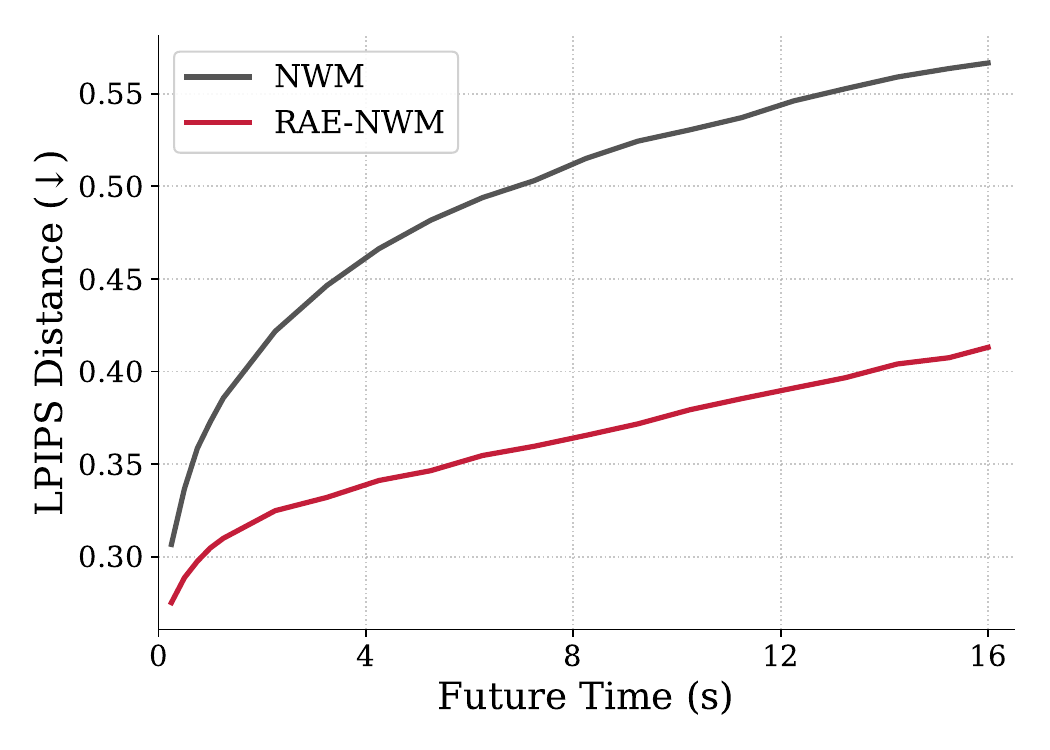}
        \caption{LPIPS}
        \label{fig:metrics_lpips}
    \end{subfigure}
    \hfill
    \begin{subfigure}[b]{0.24\textwidth}
        \centering
        \includegraphics[width=\textwidth]{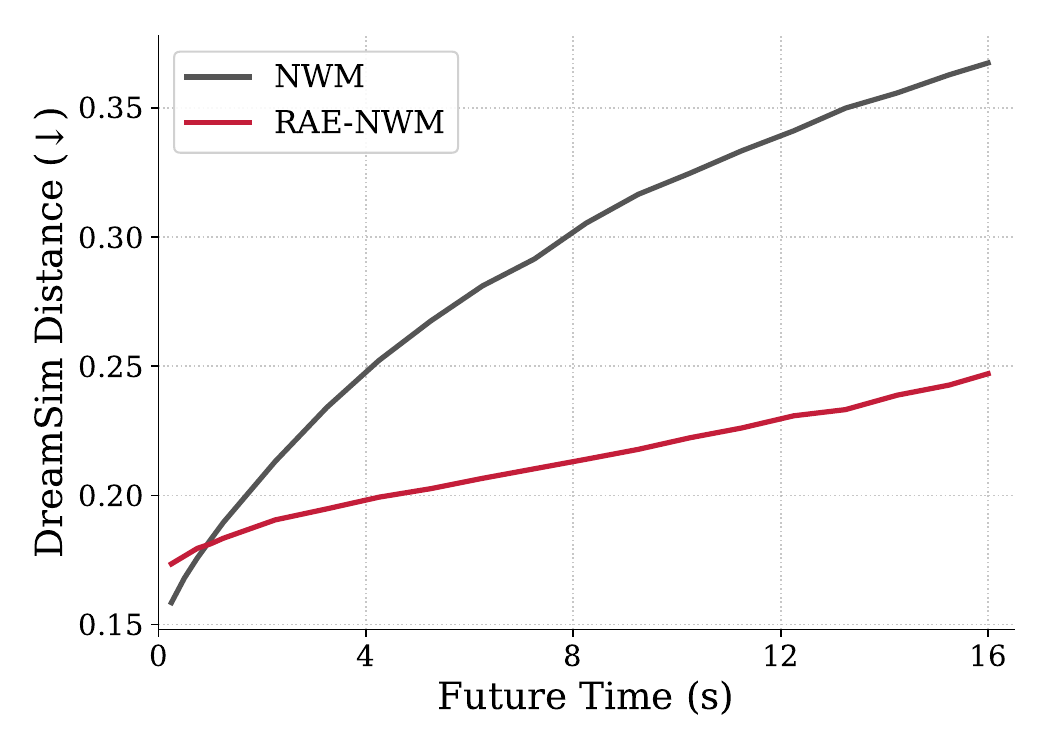}
        \caption{DreamSim}
        \label{fig:metrics_dreamsim}
    \end{subfigure}
    \hfill
    \begin{subfigure}[b]{0.24\textwidth}
        \centering
        \includegraphics[width=\textwidth]{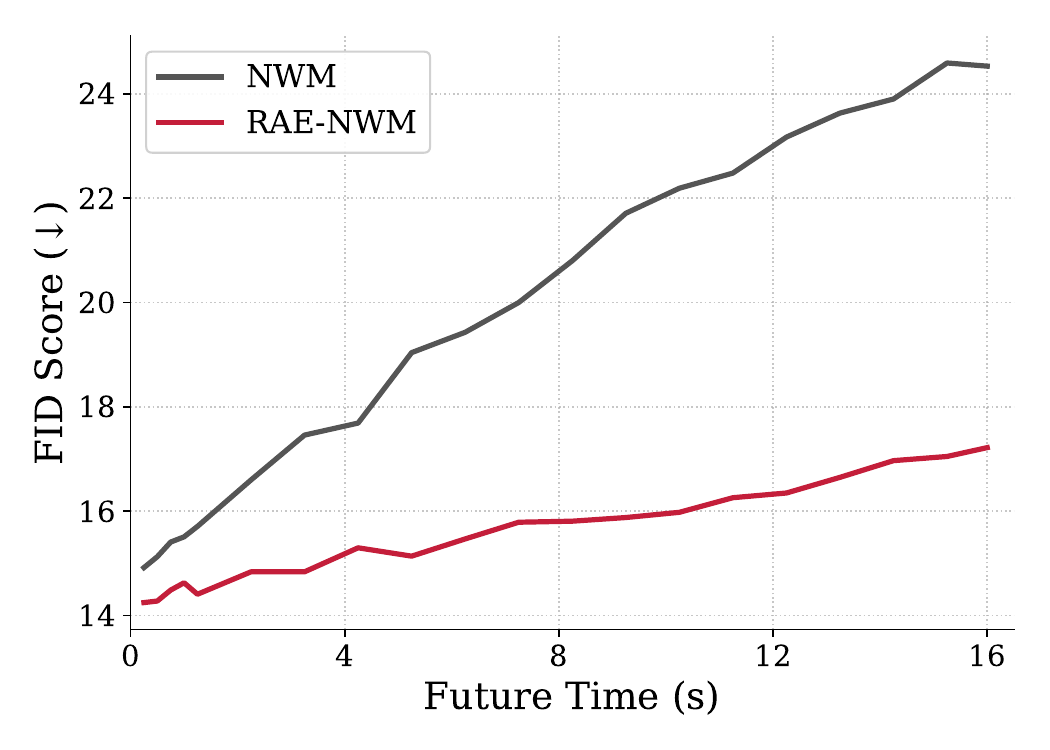}
        \caption{FID}
        \label{fig:metrics_fid}
    \end{subfigure}
    \hfill
    \begin{subfigure}[b]{0.24\textwidth}
        \centering
        \includegraphics[width=\textwidth]{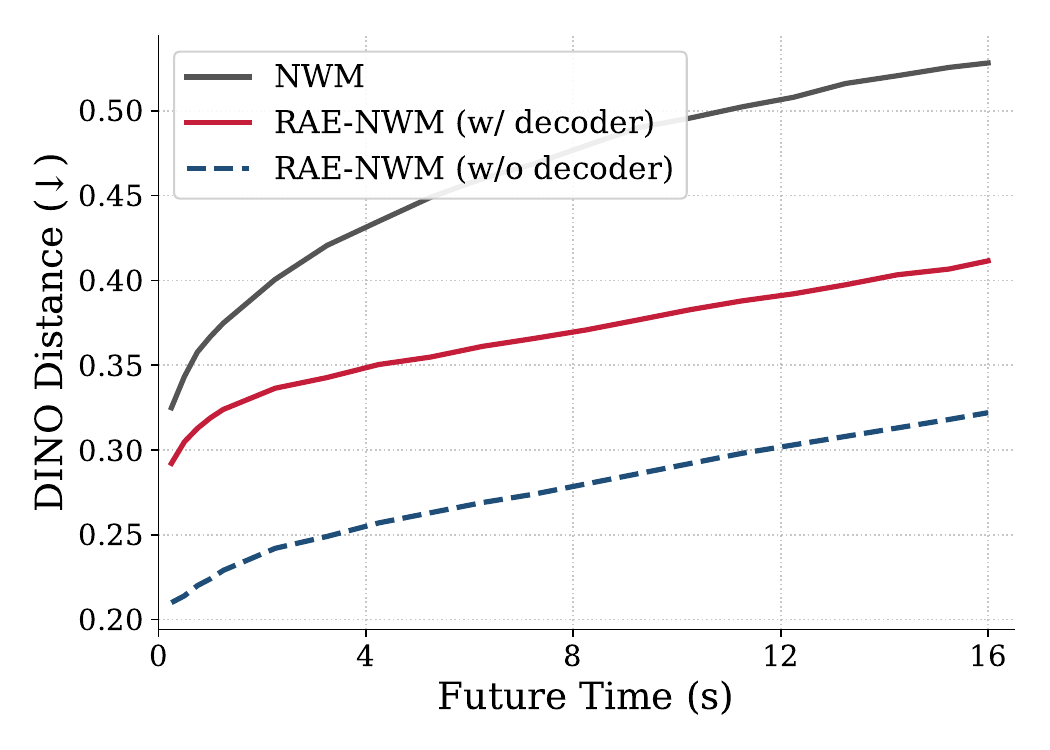}
        \caption{DINO Distance}
        \label{fig:metrics_dino}
    \end{subfigure}
    \caption{Long-horizon sequential rollout quality. Performance comparison between RAE-NWM and the baseline Navigation World Model on the SACSoN dataset over a 16-second future horizon, generated iteratively step-by-step.}
    \label{fig:all_metrics_comparison}
\end{figure*}

Building upon this direct controllability, we further evaluate open-loop generation quality using sequential rollout predictions. Given initial observations and ground-truth action sequences, the model iteratively generates 16-second future trajectories at 4 FPS on unseen test data. 
Figure~\ref{fig:all_metrics_comparison} shows the temporal evaluation on the SACSoN dataset, alongside the qualitative example provided earlier in Figure~\ref{fig:intro_compare}. Figure~\ref{fig:qualitative}b expands this qualitative evaluation to sequential rollouts on the RECON and SCAND datasets. Both the continuous rollout curves and these visualizations demonstrate that RAE-NWM maintains strong temporal stability and structural consistency over extended horizons. We further inspect rollouts beyond the 16-second horizon and observe smooth LPIPS increases from 16s to 32s on SACSoN, RECON, and SCAND: 0.387$\rightarrow$0.438, 0.458$\rightarrow$0.517, and 0.523$\rightarrow$0.614, respectively, indicating gradual degradation rather than abrupt collapse. The larger increase on SCAND suggests that moving-agent scenes remain more challenging for long-horizon rollout. To rule out the possibility that this performance is an artifact of the pre-trained RAE decoder, we compute the DINO patch distance directly in the token space against the ground truth (Figure~\ref{fig:metrics_dino}). The lower error in this uncompressed representation space confirms the capacity of the model to produce accurate structural predictions.

\begin{table}[tb]
  \caption{Trajectory prediction performance and deterministic DINO-token regression ablation. Absolute Trajectory Error (ATE) and Relative Pose Error (RPE) results are reported across all in-domain datasets for trajectories predicted up to 2 seconds. DINO-Reg denotes deterministic DINO-token regression under the same navigation protocol.}
  \label{tab:planning_accuracy}
  \centering
  \begin{tabular}{@{}lcccccc@{}}
    \toprule
    \multirow{2}{*}{Model} & \multicolumn{2}{c}{SACSoN} & \multicolumn{2}{c}{RECON} & \multicolumn{2}{c}{SCAND} \\
    \cmidrule(lr){2-3} \cmidrule(lr){4-5} \cmidrule(lr){6-7}
    & ATE $\downarrow$ & RPE $\downarrow$ & ATE $\downarrow$ & RPE $\downarrow$ & ATE $\downarrow$ & RPE $\downarrow$ \\
    \midrule
    GNM~\cite{shah2023gnm}      & 3.71 & 1.00 & 1.87 & 0.73 & 2.12 & 0.61 \\
    NoMaD~\cite{sridhar2024nomad} & 3.73 & 0.96 & 1.95 & 0.53 & 2.24 & 0.49 \\
    NWM~\cite{bar2025_nwm}      & 4.12 & 0.96 & \textbf{1.13} & \textbf{0.35} & 1.28 & 0.33 \\
    \midrule
    DINO-Reg      & 3.14 & 0.75 & 1.46 & 0.36 & 1.44 & 0.33 \\
    RAE-NWM & \textbf{2.91} & \textbf{0.70} & 1.36 & 0.37 & \textbf{1.14} & \textbf{0.28} \\
    \bottomrule
  \end{tabular}
\end{table}

\subsection{Trajectory and Planning Accuracy}
\label{subsec:planning_accuracy}

To evaluate the standalone goal-conditioned planning performance of RAE-NWM, we use the Cross-Entropy Method (CEM) for trajectory optimization. During each CEM iteration, we first sample 120 candidate action sequences. For each candidate, the world model performs an eight-step rollout in the uncompressed visual token space conditioned on the historical observations. Instead of decoding to pixel space and using low-level metrics, we compute the DINO feature distance directly in token space between the predicted rollout and the goal image. We select the highest-scoring trajectory for execution. To quantify planning accuracy, we compute the Absolute Trajectory Error (ATE) and Relative Pose Error (RPE) against the ground-truth trajectory. Table~\ref{tab:planning_accuracy} also includes a deterministic DINO-token regression variant, which is further discussed in the ablation study.

Table~\ref{tab:planning_accuracy} shows that RAE-NWM improves upon the primary baseline NWM on the SACSoN dataset, reducing the ATE to 2.91 and RPE to 0.70. This trend continues on the SCAND dataset. On the unstructured RECON dataset, the proposed model yields a competitive ATE of 1.36 and RPE of 0.37, outperforming end-to-end policies like GNM and NoMaD. High-frequency visual elements, such as the dense off-road vegetation in the RECON dataset, introduce stochastic textures that complicate environment modeling. In this short-horizon setting, NWM obtains lower ATE and RPE on RECON. This performance gap likely occurs because the VAE-based latent space preserves more texture-sensitive cues, while DINOv2 features emphasize semantic and structural information.

\subsection{Simulation in Habitat}
\label{subsec:habitat_simulation}

To evaluate closed-loop control capabilities, we deploy RAE-NWM as a dynamics engine in the Habitat simulator using scenes from Matterport3D. We assess the model on the Image-Goal navigation task, where the agent navigates to a destination specified by a goal image. An episode (average length 8 meters) is considered successful if the agent stops within 1 meter of the target. At each environment step, the agent performs CEM-based trajectory optimization guided by the same token-space DINO feature distance used in Section~\ref{subsec:planning_accuracy}, and executes the selected action accordingly. Table~\ref{tab:habitat_results} shows that RAE-NWM achieves a Success Rate of 78.95\%, outperforming existing visual navigation methods. Its SPL is slightly lower than that of One-Step WM, likely because one-step generation can evaluate longer planning trajectories more efficiently under a fixed computation budget, while RAE-NWM attains a higher SR with iterative dense-space rollouts.

\begin{table}[tb]
  \begin{minipage}[t]{0.48\linewidth}
    \centering
    \caption{Closed-loop image-goal navigation results in Habitat.}
    \label{tab:habitat_results}
    \begin{tabular}{@{}lcc@{}}
      \toprule
      Method & SR (\%) $\uparrow$ & SPL (\%) $\uparrow$ \\
      \midrule
      NoMaD~\cite{sridhar2024nomad}             & 16.67 & 15.13 \\
      OmniVLA~\cite{hirose2025omnivla}           & 36.67 & 34.78 \\
      NWM~\cite{bar2025_nwm}             & 43.33 & 38.66 \\
      One-Step WM~\cite{shen2026efficient}        & 72.67 & \textbf{69.10} \\
      \midrule
      RAE-NWM                 & \textbf{78.95} & 63.58 \\
      \bottomrule
    \end{tabular}
  \end{minipage}\hfill
  \begin{minipage}[t]{0.48\linewidth}
    \centering
    \caption{Dynamics condition ablation studies on the SACSoN dataset.}
    \label{tab:ablation_plan}
    \begin{tabular}{@{}lcc@{}}
      \toprule
      Method & ATE $\downarrow$ & RPE $\downarrow$ \\
      \midrule
      Simple Addition & 3.34 & 0.79 \\
      MLP Fusion & 3.54 & 0.83 \\
      Scheduled Gate & 3.31 & 0.78 \\
      Learned Gate (ours) & \textbf{2.91} & \textbf{0.70} \\
      \bottomrule
    \end{tabular}
  \end{minipage}
\end{table}

\subsection{Ablation Study}
\label{subsec:ablation}

To study the impact of each component, we evaluate both long-horizon rollout quality and downstream planning accuracy. Sequential generation over a 16-second horizon serves as a stress test for temporal stability, while the planning task further measures action controllability.

\begin{figure*}[tb] 
  \centering
  \begin{minipage}{0.48\textwidth}
    \centering
    \includegraphics[width=\linewidth]{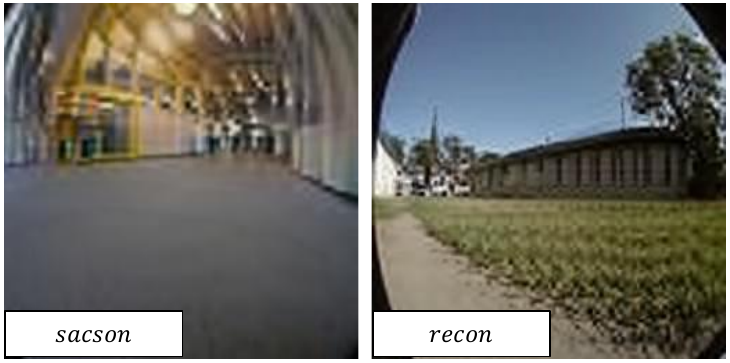}
    \caption{Qualitative rollout visualization of DINO-Reg.}
    \label{fig:dino_reg_vis}
  \end{minipage}
  \hfill
  \begin{minipage}{0.48\textwidth}
    \centering
    \includegraphics[width=0.49\linewidth]{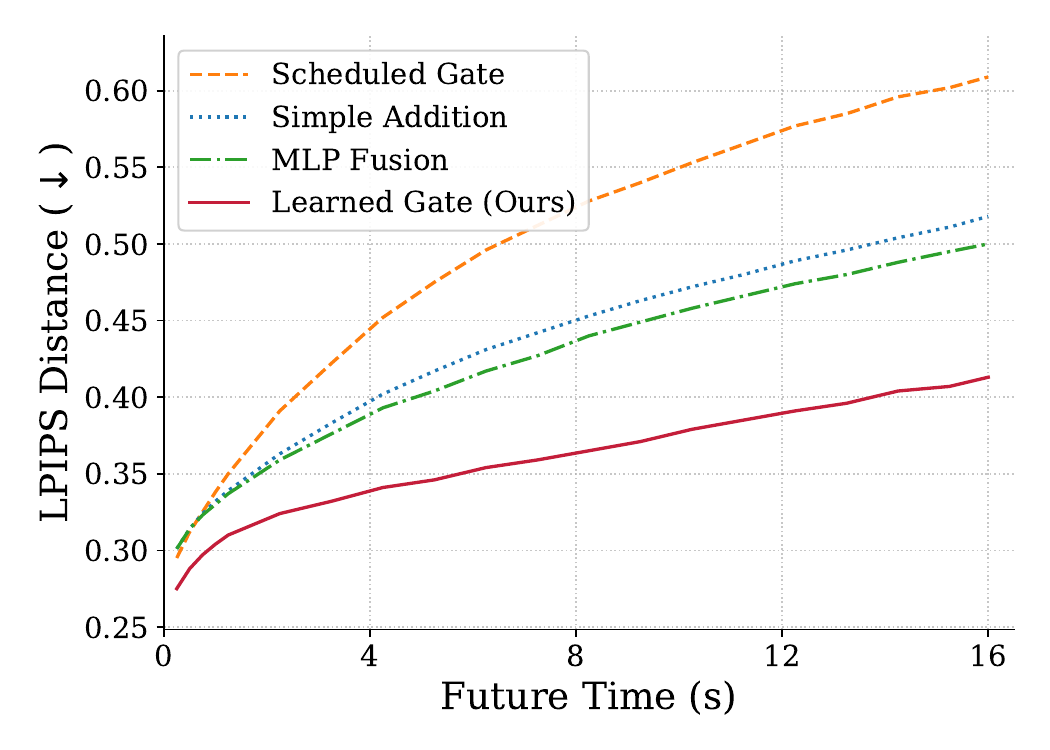}
    \hfill
    \includegraphics[width=0.49\linewidth]{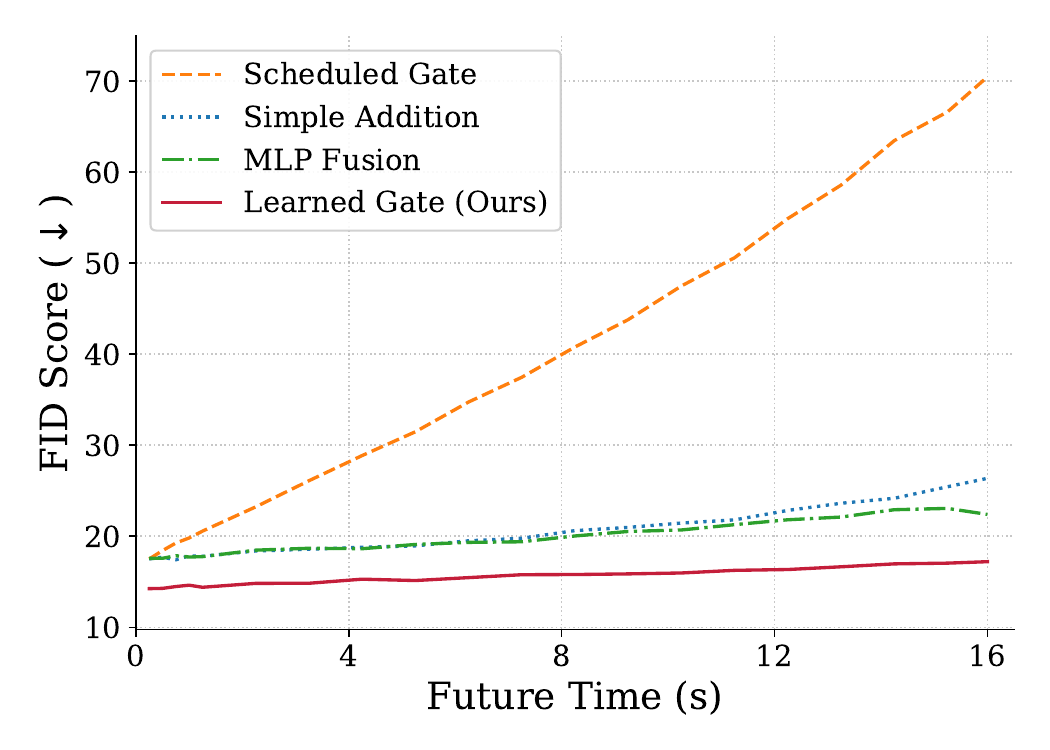}
    \caption{Ablation on dynamics condition injection strategies.}
    \label{fig:ablation_action}
  \end{minipage}
\end{figure*}

\paragraph{Transition Modeling Objective.}
To isolate the effect of the generative transition formulation from the choice of dense DINO representation, we implement a deterministic DINO-token Regression (DINO-Reg) variant under the same first-person navigation protocol. This variant keeps the representation space and evaluation protocol unchanged, but replaces flow matching with direct MSE regression over future DINO tokens. As shown in Table~\ref{tab:planning_accuracy}, RAE-NWM obtains lower ATE on all datasets and lower RPE on SACSoN and SCAND, while DINO-Reg is marginally better on RECON RPE (0.36 vs. 0.37). This suggests that the transition objective contributes beyond the representation choice alone. The qualitative rollout in Figure~\ref{fig:dino_reg_vis} further shows that deterministic regression produces blurry predictions with ghosting artifacts, rather than stable frame-wise rollouts.

\paragraph{Dynamics Conditioning Module.}
To investigate the optimal condition injection strategy during the continuous generative process, we compare our adaptive learned gate against simple addition, MLP fusion, and scheduled gate. As Figure~\ref{fig:ablation_action} illustrates, standard injection strategies and the scheduled gate experience continuous degradation in both LPIPS and FID over extended horizons, indicating that rigid conditioning is less effective at preserving spatial structure. In contrast, the learned gate maintains the lowest LPIPS trajectory and a more stable FID curve. We further evaluate these variants in downstream planning on SACSoN (Table~\ref{tab:ablation_plan}), where the learned gate achieves the lowest ATE of 2.91 and RPE of 0.70. These results show that adaptively modulating the injected dynamics representation reduces visual error accumulation and improves precise action controllability. Detailed conditioning designs and gate dynamics are provided in Section D of the Supplementary Material.

\paragraph{Representation Space and Architecture.}
To test the impact of the representation space, we substitute the DINOv2 encoder with the same frozen SD-VAE~\cite{rombach2022high} used in NWM. As illustrated in Figure~\ref{fig:ablation_encoder}, while SD-VAE maintains a lower initial LPIPS due to its pixel-level reconstruction objective, it suffers from stronger degradation over extended horizons. The DINOv2-based model quickly surpasses it, exhibiting better long-term spatial stability and a superior final FID. We also remove the DDT head to evaluate the architectural design. Without the DDT head, the long-horizon LPIPS exceeds the SD-VAE baseline, suggesting that directly modeling high-dimensional DINOv2 representations remains difficult. Consistent with RAE~\cite{zheng2025rae}, the DDT head effectively improves optimization in this dense token space, enabling the model to better exploit the geometric priors of DINOv2 during continuous rollouts.

\section{Discussion and Limitations}
\label{sec:discussion}

\begin{figure*}[tb] 
  \centering
  \begin{minipage}{0.48\textwidth}
    \centering
    \includegraphics[width=0.49\linewidth]{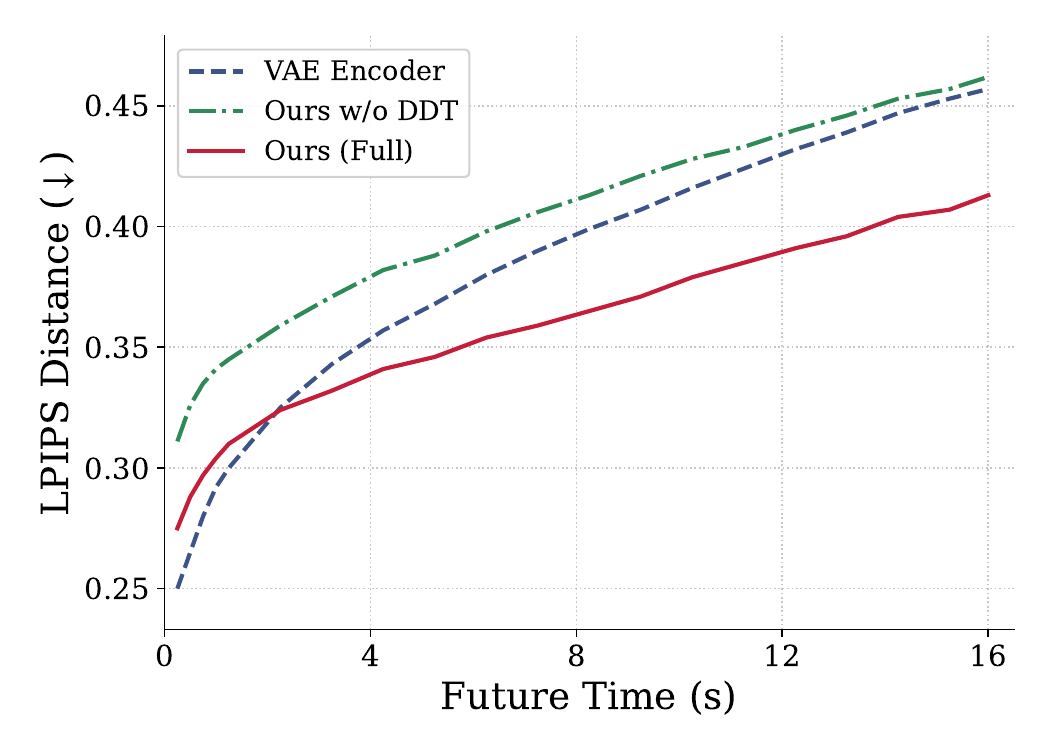} 
    \hfill
    \includegraphics[width=0.49\linewidth]{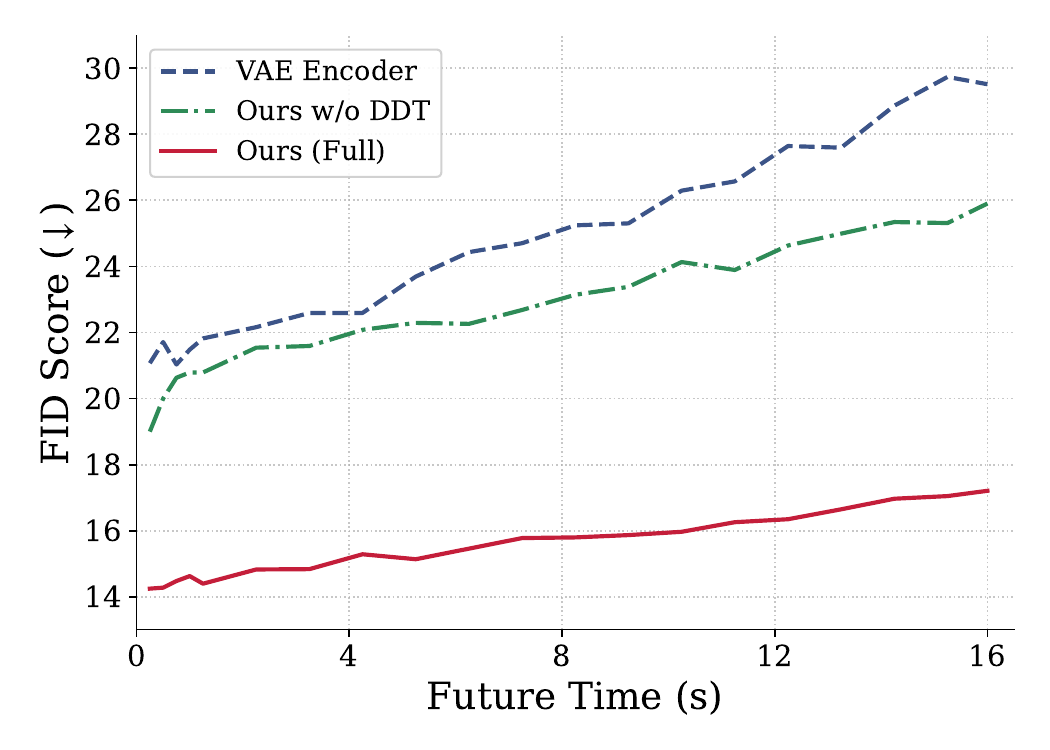}
    \caption{Ablations on visual encoders and the DDT head.}
    \label{fig:ablation_encoder}
  \end{minipage}
  \hfill
  \begin{minipage}{0.48\textwidth}
    \centering
    \includegraphics[width=0.49\linewidth]{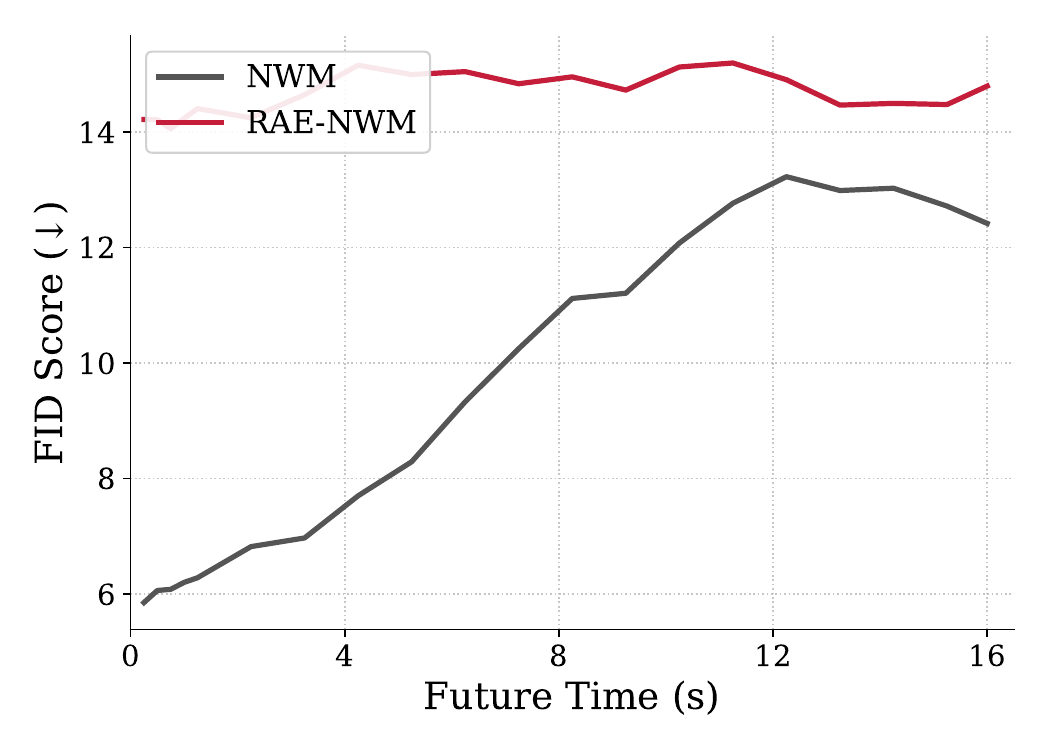}
    \hfill
    \includegraphics[width=0.49\linewidth]{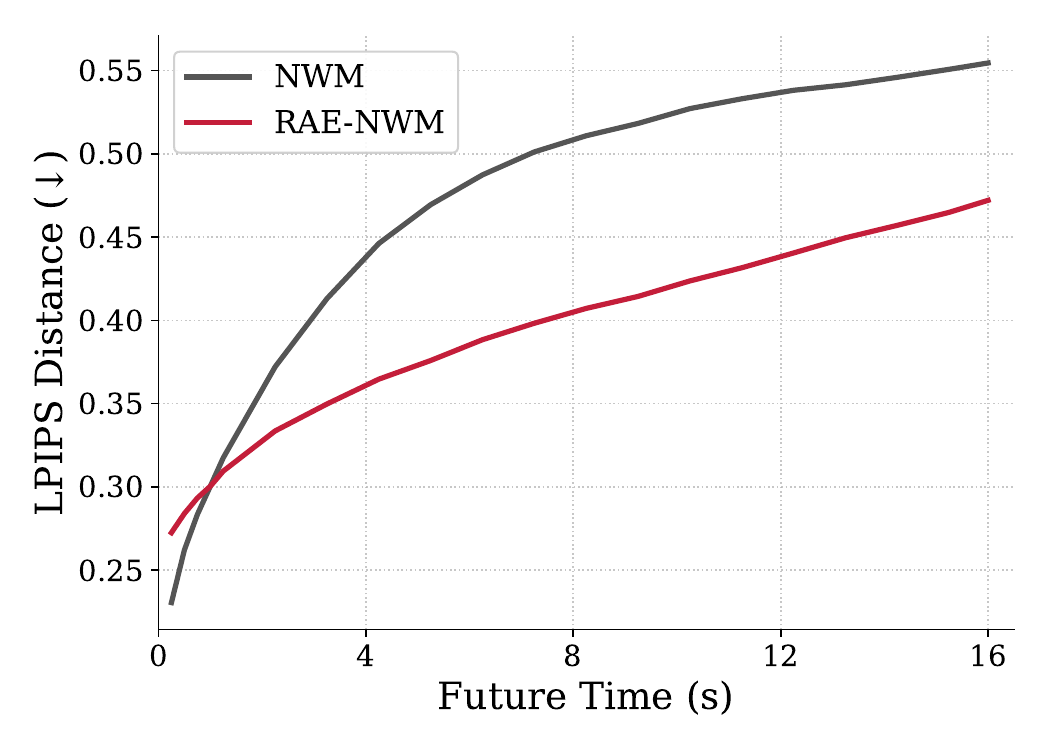}
    \caption{Quantitative performance on the RECON dataset.}
    \label{fig:recon_discussion}
  \end{minipage}
\end{figure*}

While RAE-NWM excels in structured environments, semantic representations like DINOv2 tend to overlook high-frequency stochastic textures (e.g., grass in the RECON dataset). Figure~\ref{fig:recon_discussion} illustrates that this initial fidelity loss acts as a trade-off for long-horizon spatial stability. The NWM starts with a slightly lower LPIPS (0.230), but its limited structural grounding leads to rapid degradation, reaching 0.554 during sequential rollouts. RAE-NWM maintains a more stable trajectory, ending at 0.472. This indicates that semantic priors help maintain coherent geometric structure during dynamics generation, even though some texture-sensitive details are less faithfully preserved. We further measure static encode-decode fidelity on RECON without dynamics prediction, where SD-VAE and DINOv2+RAE obtain FID scores of 4.14 and 5.47, respectively, confirming the texture-fidelity cost of DINOv2+RAE.

Despite using a smaller generative backbone than NWM (350M vs. 1B parameters; 256 vs. 784 dynamics tokens), RAE-NWM achieves improved performance with lower dynamics-forward cost in our profiling (8.97 vs. 36.10 TFLOPs; 19.4 vs. 28.8 ms; 0.73 vs. 1.93 GB on one A800). Fully scale-matched training remains a limitation for isolating the effects of representation choice, architecture, and model capacity.

\section{Conclusion}

We present the Representation Autoencoder-based Navigation World Model (RAE-NWM), a generative navigation world model that learns action-conditioned transitions in the dense DINOv2 representation space. Built upon a frozen DINOv2 encoder and a frozen RAE decoder, RAE-NWM trains a CDiT-DH backbone to model continuous transition dynamics and introduces a time-driven gating mechanism to adaptively modulate kinematic conditioning along the probability flow. This design improves long-horizon rollout stability, preserves spatial geometric structure, and provides more reliable predictions for downstream planning. Extensive experiments on real-world navigation datasets and Habitat Image-Goal navigation demonstrate the effectiveness of dense representation-space world modeling. Future work will explore improving texture fidelity and accelerating iterative rollouts for more efficient planning.

\section*{Acknowledgments}

This work was supported in part by Tsinghua-Toyota Joint Research Fund, in part by National Natural Science Foundation of China (Grant No. 62403269), and in part by Beijing Natural Science Foundation under grant number L233029.

\bibliographystyle{splncs04}
\bibliography{main}

% ==========================================
% Supplementary Material starts here
% ==========================================
\clearpage
\appendix

\begin{center}
    \Large \textbf{Supplementary Material}
\end{center}

\section{Implementation Details}
\label{supp:implementation}

\subsection{Network Architecture}
The proposed CDiT-DH backbone follows the DiT-B configuration, utilizing 12 Transformer blocks with a hidden size of 768 and 12 attention heads. The DDT head is designed to be shallow and wide, comprising 2 blocks with a hidden size of 2048 and 16 attention heads. For the visual input, we employ a context window of 4 context frames.

\subsection{Training and Optimization}
The models are trained for a total of 50 epochs. We use the AdamW optimizer with a total batch size of 96. The learning rate is initialized at $2\times10^{-4}$ and linearly decayed to $2\times10^{-6}$ over the course of training. The weight decay is set to 0.

\subsection{Inference and Training Resources}
During inference, we employ an Euler ordinary differential equation (ODE) solver with 50 integration steps. For the real-world experiments, we train one unified model jointly on SACSoN, RECON, and SCAND using two NVIDIA A800 80GB GPUs, and training takes approximately 2 days.

\subsection{Action Normalization}
To align the kinematic scales across different robotic platforms, we normalize the action magnitude by the dataset-specific average inter-frame step size computed from the training split of each dataset. This normalization ensures that actions from different datasets correspond to comparable physical motion scales and reduces dataset-specific scale discrepancies during joint training.

\subsection{DINO Representation Distance}
To measure similarity in the visual representation space, we compute a DINO feature distance based on cosine similarity between patch tokens.

Given two raw visual observations $\mathbf{o}_1$ and $\mathbf{o}_2$, we extract their patch token features using a frozen DINOv2 encoder:
\[
\mathbf{z}_1, \mathbf{z}_2 \in \mathbb{R}^{L \times d},
\]
where $L$ denotes the number of patch tokens and $d$ is the feature dimension.

The features are first $\ell_2$ normalized along the channel dimension. For each token index $l$, the cosine distance is computed as
\[
d_l = 1 - \langle \mathbf{z}_{1,l}, \mathbf{z}_{2,l} \rangle .
\]

The final DINO distance between the two observations is obtained by averaging over all tokens:
\[
\mathrm{DINO}(\mathbf{o}_1, \mathbf{o}_2) = \frac{1}{L}\sum_{l=1}^{L} d_l .
\]

\section{Technical Details of the Linear Dynamics Probe}
\label{appendix:probe_details}

This section provides implementation details, training configuration, and metric computation for the linear dynamics probe used in the representation analysis.

\subsection{Implementation of Linear Transformations}

The probe consists of two learnable linear transformations, $\mathbf{A}$ and $\mathbf{B}$. To avoid introducing additional modeling capacity, both transformations are implemented as linear operators without non-linear activation functions.

The spatial transformation $\mathbf{A}$ is implemented as a $1 \times 1$ convolution without bias. The visual representation is first obtained as a token sequence $\mathbf{z}_i \in \mathbb{R}^{L \times d}$ and then reshaped into a feature map $\tilde{\mathbf{z}}_i \in \mathbb{R}^{d \times H_p \times W_p}$, where $H_p$ and $W_p$ denote the patch-grid height and width, and $L = H_p W_p$. The $1 \times 1$ convolution operates across the channel dimension of this reshaped feature map.

The action transformation $\mathbf{B}$ processes the motion condition $\bm{a}_{i \rightarrow i+k} \in \mathbb{R}^3$, which contains the relative planar translation and yaw rotation $(u_x, u_y, \omega)$. This transformation is implemented as a bias-free linear layer projecting the input action vector to the feature dimension $d$. The resulting vector is spatially broadcast to match the resolution $H_p \times W_p$ before being added to the spatial features.

\subsection{Training Configuration and Dataset Specifics}

To ensure a fair comparison between representation spaces, the trajectories used to train the probe are separated from those used for evaluation. During probe training, the parameters of all visual encoders remain frozen.

The probe is optimized using the Huber loss between the predicted and ground-truth latent states. The temporal step size $k$ corresponds to a non-trivial physical displacement to ensure that the task requires meaningful kinematic prediction. The average metric displacement per frame is approximately 0.25\,m for RECON, 0.255\,m for SACSoN, and 0.36\,m for SCAND. We optimize the linear probe using the AdamW optimizer with a weight decay of $10^{-4}$, where the learning rate follows a linear warmup and cosine decay schedule.

\subsection{Calculation of the Global $R^2$ Score}

To compute the global $R^2$ score reported in the representation analysis, the predicted latent state $\hat{\mathbf{z}}_{i+k}$ and the ground-truth latent state $\mathbf{z}_{i+k}$ are flattened across the batch, spatial, and channel dimensions into one-dimensional vectors, denoted as $\hat{\mathbf{y}}$ and $\mathbf{y}$.

We compute the Sum of Squared Errors (SSE) and the Total Sum of Squares (SST) as
\[
\mathrm{SSE} = \sum_{j} (y_{j} - \hat{y}_{j})^2 ,
\]
\[
\mathrm{SST} = \sum_{j} (y_{j} - \bar{y})^2 ,
\]
where $\bar{y}$ denotes the mean of the ground-truth vector $\mathbf{y}$. The global coefficient of determination is then computed as
\[
R^2 = 1 - \frac{\mathrm{SSE}}{\mathrm{SST}} .
\]

\section{Details of Deterministic DINO-Token Regression}
\label{supp:dino_reg}

To further isolate the effect of the transition modeling objective, we implement a deterministic DINO-token regression baseline under the same navigation rollout protocol. This variant uses the same frozen DINOv2 representation space, context inputs, action conditions, and rollout evaluation protocol as RAE-NWM, but replaces the flow matching objective with direct regression to the future DINO tokens:
\[
\mathcal{L}_{\mathrm{reg}}
=
\left\|
\mathrm{Pred}_{\theta}
\left(
\mathbf{z}_{\mathrm{cond},i},
\bm{a}_{i\rightarrow i+k},
k
\right)
-
\mathbf{z}_{i+k}
\right\|_2^2 .
\]
This baseline follows the deterministic feature-prediction formulation of DINO-space predictive world models while avoiding changes to the navigation protocol, decoder, and evaluation setting. In the trajectory evaluation, deterministic regression improves over several navigation baselines, but the proposed flow-matching formulation achieves stronger overall performance and more stable rollouts.

\section{Extended Details on the Dynamics Conditioning Module}
\label{supp:dynamics_module}

In this section, we provide detailed architectural comparisons of the condition injection strategies discussed in the main paper and further analyze the learned behavior of the proposed time-driven dynamic gating mechanism.

\begin{figure*}[tbp]
    \centering
    \begin{subfigure}[b]{0.24\textwidth}
        \centering
        \includegraphics[width=\textwidth]{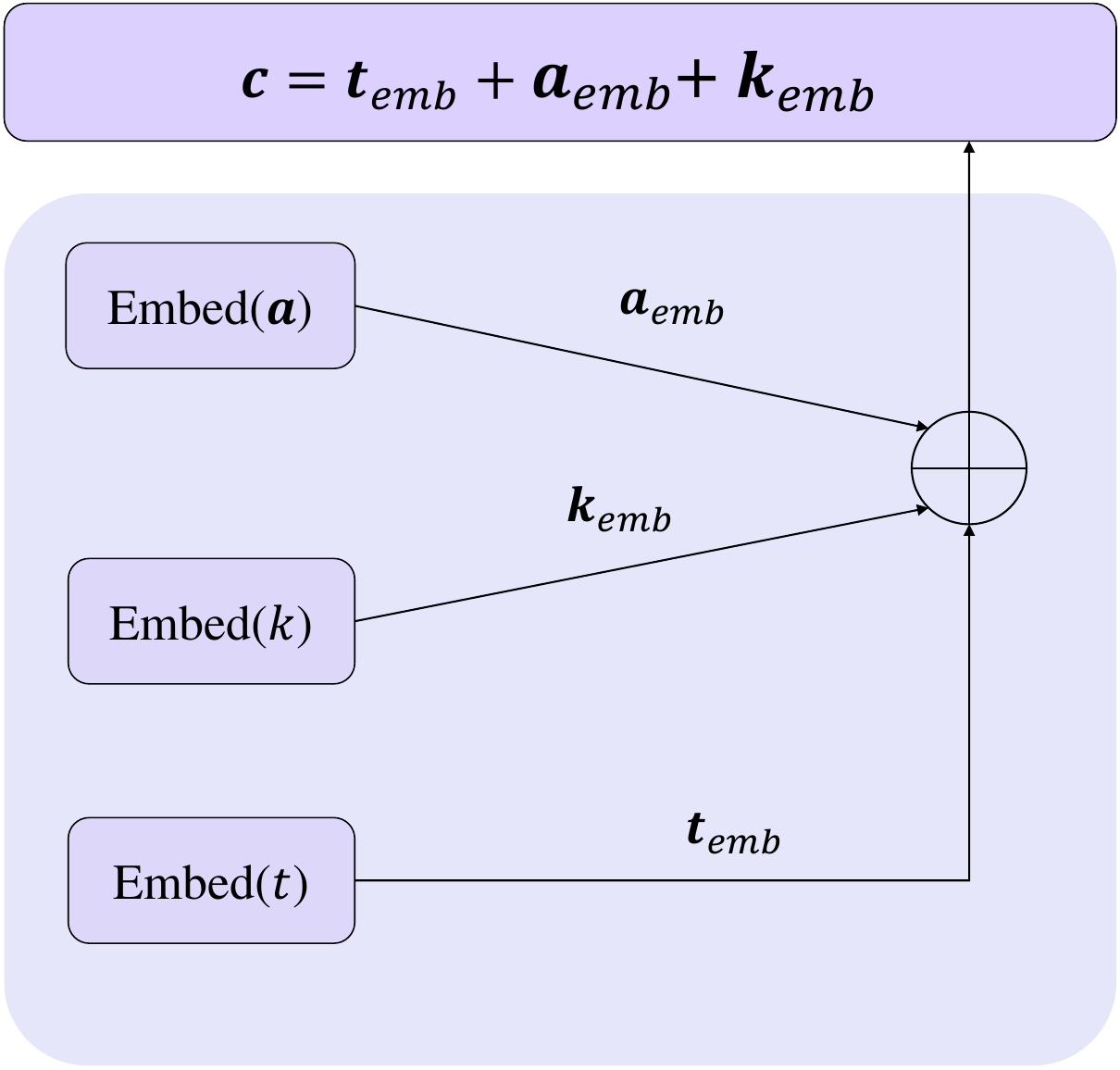}
        \caption{Simple Addition}
        \label{fig:ablation_simple}
    \end{subfigure}
    \hfill
    \begin{subfigure}[b]{0.24\textwidth}
        \centering
        \includegraphics[width=\textwidth]{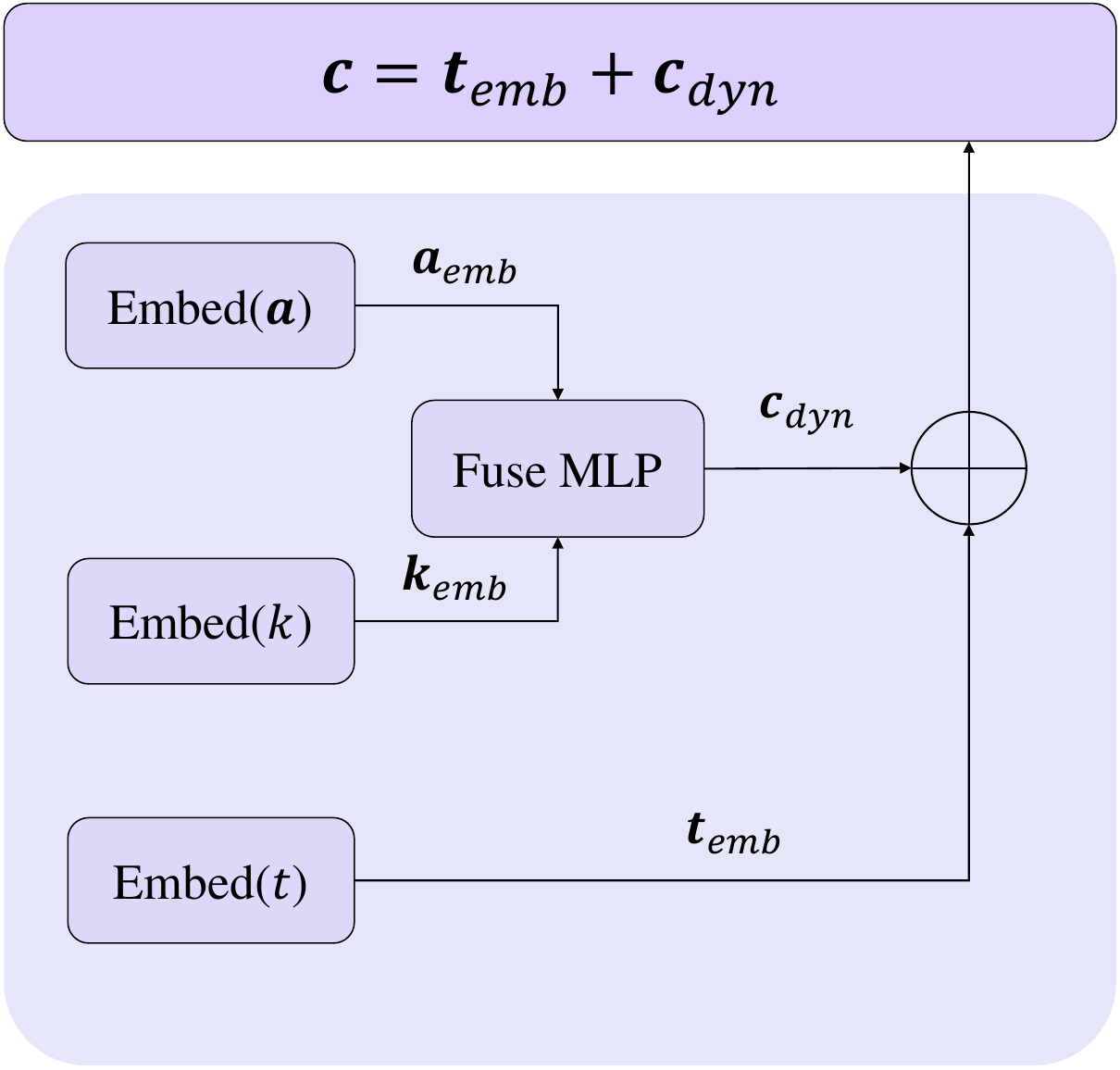}
        \caption{MLP Fusion}
        \label{fig:ablation_mlp}
    \end{subfigure}
    \hfill
    \begin{subfigure}[b]{0.24\textwidth}
        \centering
        \includegraphics[width=\textwidth]{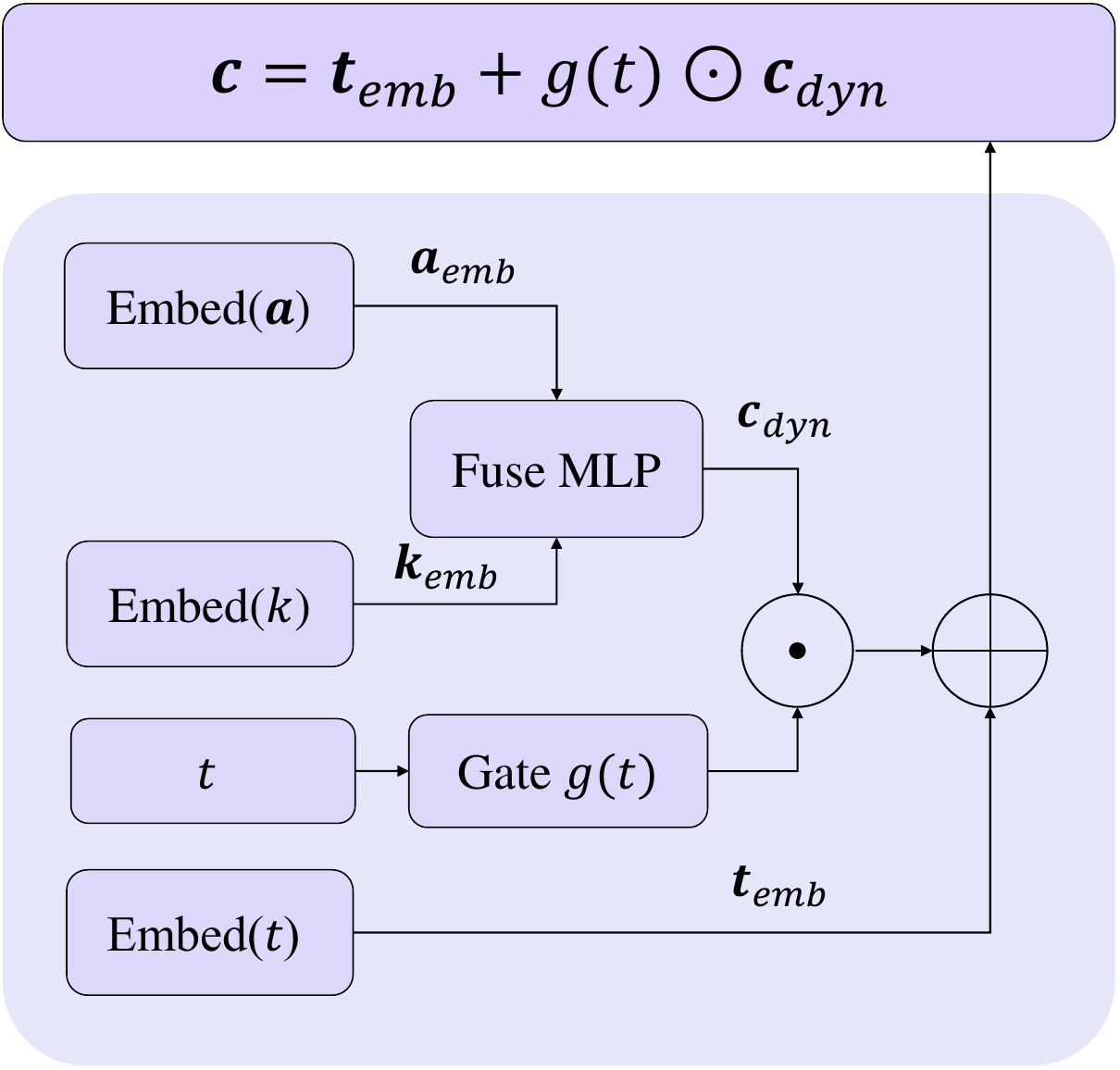}
        \caption{Scheduled Gate}
        \label{fig:ablation_scheduled}
    \end{subfigure}
    \hfill
    \begin{subfigure}[b]{0.24\textwidth}
        \centering
        \includegraphics[width=\textwidth]{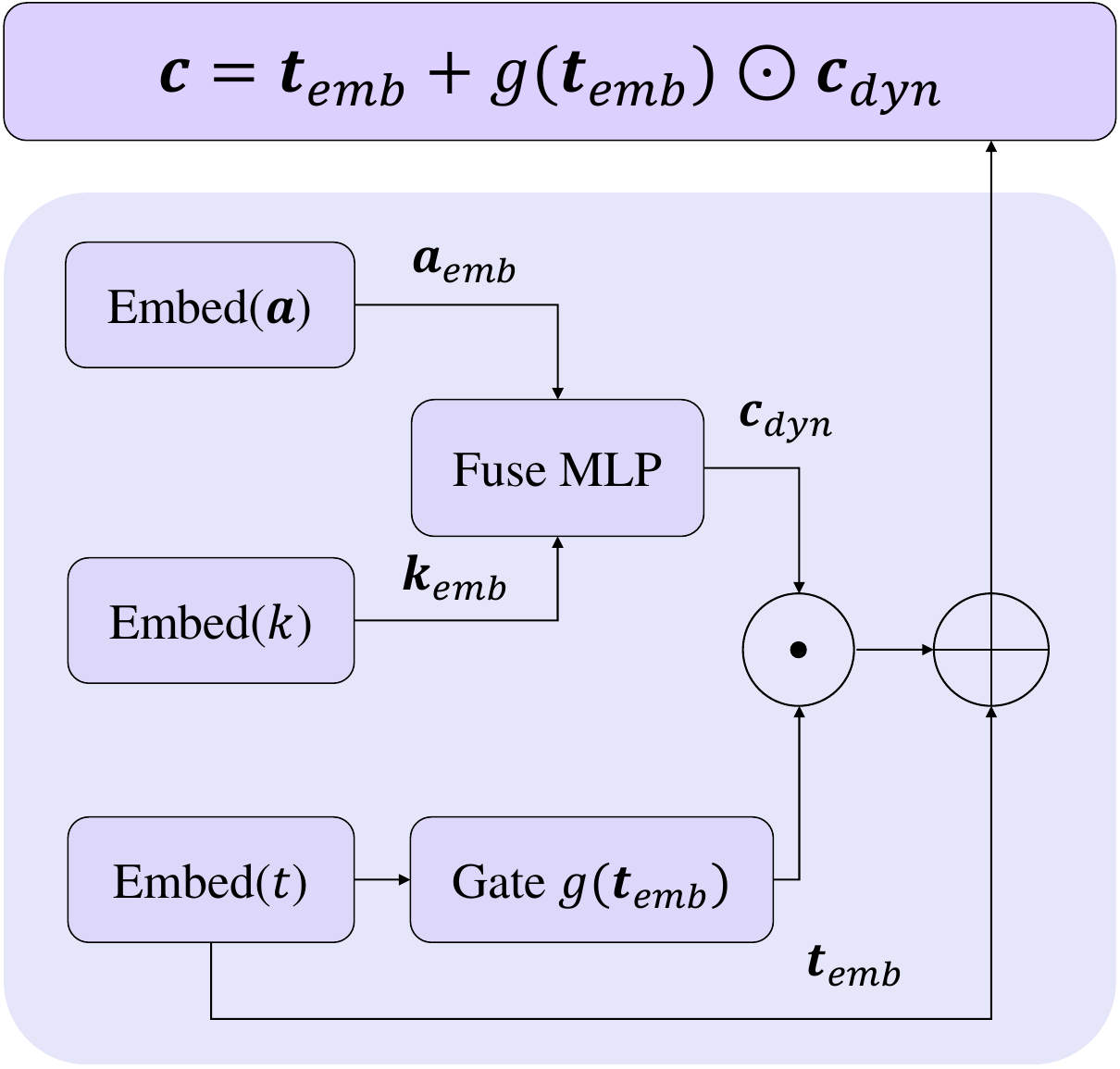}
        \caption{Learned Gate}
        \label{fig:ablation_learned}
    \end{subfigure}
    \caption{\textbf{Architectural designs for action and time-span encoding.} The figure details the four injection variants compared in the ablation study.}
    \label{fig:action_encoding_ablation}
\end{figure*}

\paragraph{Action and Time-Span Encoding Architectures.}
Detailed architectural comparisons of the condition injection variants evaluated in the ablation study are illustrated in Figure~\ref{fig:action_encoding_ablation}.

\paragraph{Analysis of Time-Driven Dynamic Gating.}
To analyze the temporal behavior of the learned gate, we evaluate the $L_2$ norm of the time embedding $\|\mathbf{t}_{\mathrm{emb}}\|_2$ and the modulated dynamics feature $\|g(\mathbf{t}_{\mathrm{emb}}) \odot \mathbf{c}_{\mathrm{dyn}}\|_2$. To make the comparison scale-invariant, we report two metrics at a given flow time $t$: the Dynamics Proportion $\mathcal{P}_{\mathrm{dyn}}$ and the Dynamics-to-Time Ratio $\mathcal{R}_{\mathrm{dyn}}$:
\[
\mathcal{P}_{\mathrm{dyn}}(t)
=
\frac{
\|g(\mathbf{t}_{\mathrm{emb}}) \odot \mathbf{c}_{\mathrm{dyn}}\|_2
}{
\|\mathbf{t}_{\mathrm{emb}}\|_2
+
\|g(\mathbf{t}_{\mathrm{emb}}) \odot \mathbf{c}_{\mathrm{dyn}}\|_2
},
\]
\[
\mathcal{R}_{\mathrm{dyn}}(t)
=
\frac{
\|g(\mathbf{t}_{\mathrm{emb}}) \odot \mathbf{c}_{\mathrm{dyn}}\|_2
}{
\|\mathbf{t}_{\mathrm{emb}}\|_2 + \epsilon
}.
\]
Here $\epsilon$ is a small constant used for numerical stability.

\begin{figure}[tbp]
    \centering
    \begin{subfigure}[b]{0.48\textwidth}
        \centering
        \includegraphics[width=\textwidth]{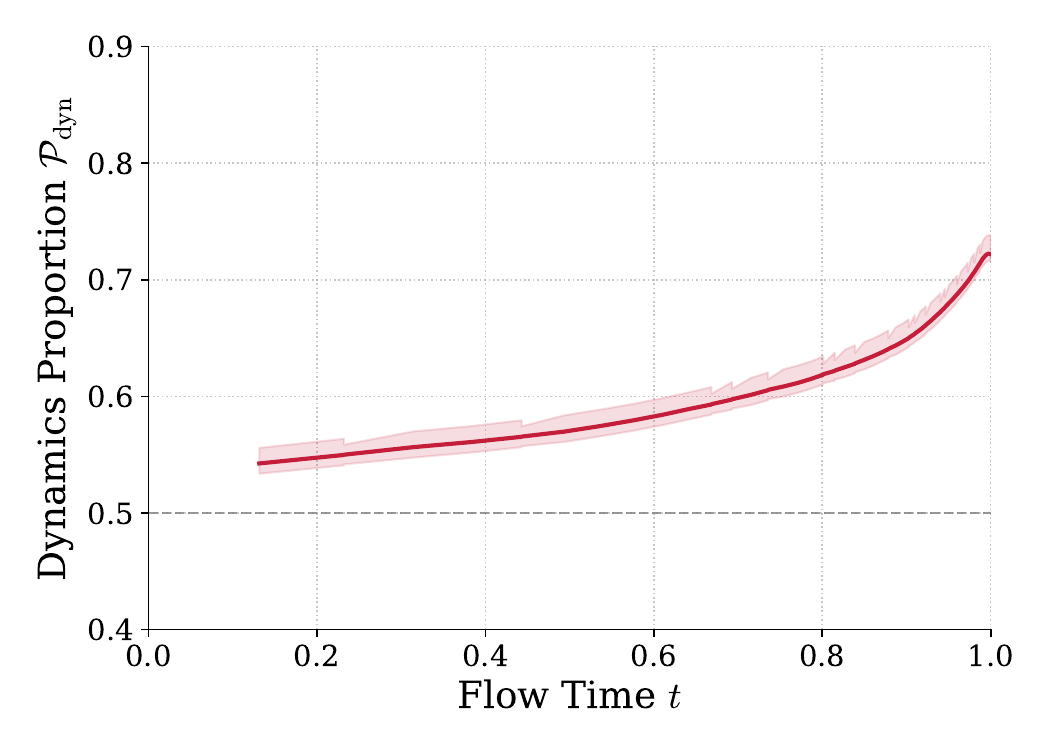}
        \caption{Dynamics Proportion $\mathcal{P}_{\mathrm{dyn}}$}
        \label{fig:gate_proportion}
    \end{subfigure}
    \hfill
    \begin{subfigure}[b]{0.48\textwidth}
        \centering
        \includegraphics[width=\textwidth]{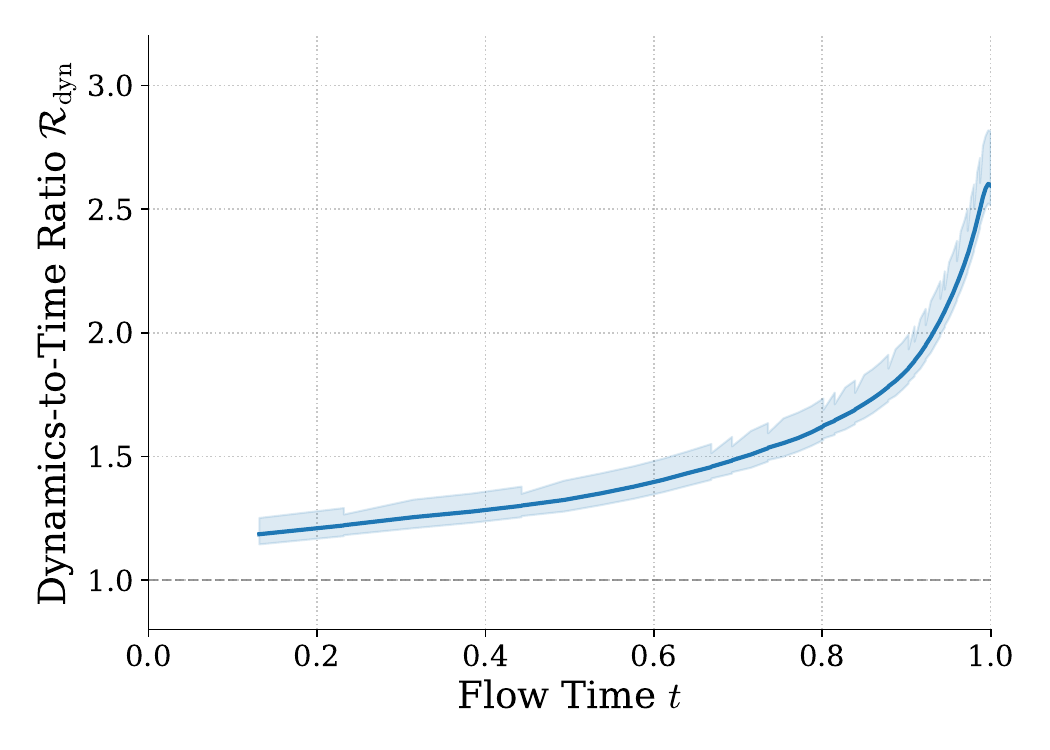}
        \caption{Dynamics-to-Time Ratio $\mathcal{R}_{\mathrm{dyn}}$}
        \label{fig:gate_ratio}
    \end{subfigure}
    \caption{\textbf{Quantitative analysis of the time-driven dynamic gating.} The curves illustrate the temporal evolution of the dynamics conditioning strength across the continuous probability flow.}
    \label{fig:gate_visualization}
\end{figure}

As visualized in Figure~\ref{fig:gate_visualization}, the network learns to maintain stronger dependence on the kinematic prior during high-noise stages ($t \to 1$), which provides structural guidance for establishing global topology. As the denoising process approaches the clean data manifold ($t \to 0$), the relative intensity of the dynamics condition decays, allowing the model to refine visual details with weaker rigid motion constraints. The narrow interquartile ranges also indicate that this temporal allocation is stable across diverse evaluation samples.

\section{Extended Evaluation on Sequential Rollouts}
\label{supp:extended_rollouts}

To provide a more comprehensive assessment of long-horizon generation across diverse environments, we expand the open-loop sequential rollout evaluation with additional quantitative metrics and qualitative examples on RECON, SCAND, SACSoN, and Matterport3D.

\subsection{Quantitative Performance}

Following the same evaluation protocol, the model iteratively generates future observations at 4 FPS based on initial context frames and ground-truth action sequences. We evaluate the generated trajectories using LPIPS, DreamSim, FID, and DINO Distance.

\begin{figure}[tbp]
    \centering
    \begin{subfigure}[b]{0.24\textwidth}
        \centering
        \includegraphics[width=\textwidth]{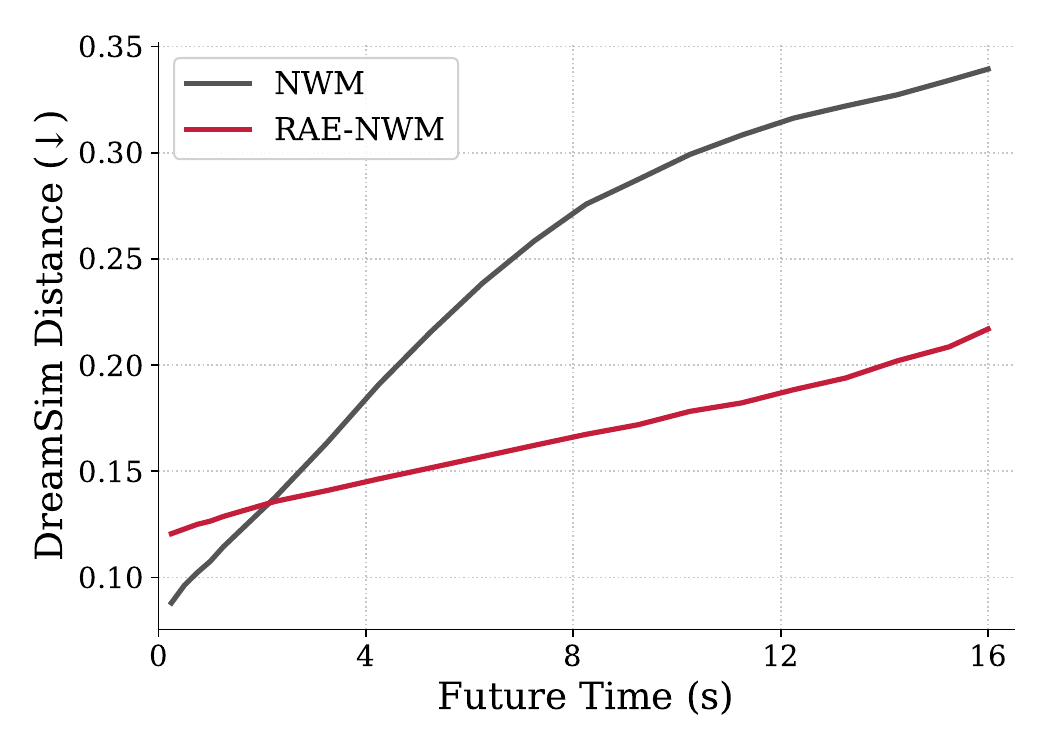}
        \caption{DreamSim}
        \label{fig:recon_dreamsim}
    \end{subfigure}
    \hspace{0.05\textwidth}
    \begin{subfigure}[b]{0.24\textwidth}
        \centering
        \includegraphics[width=\textwidth]{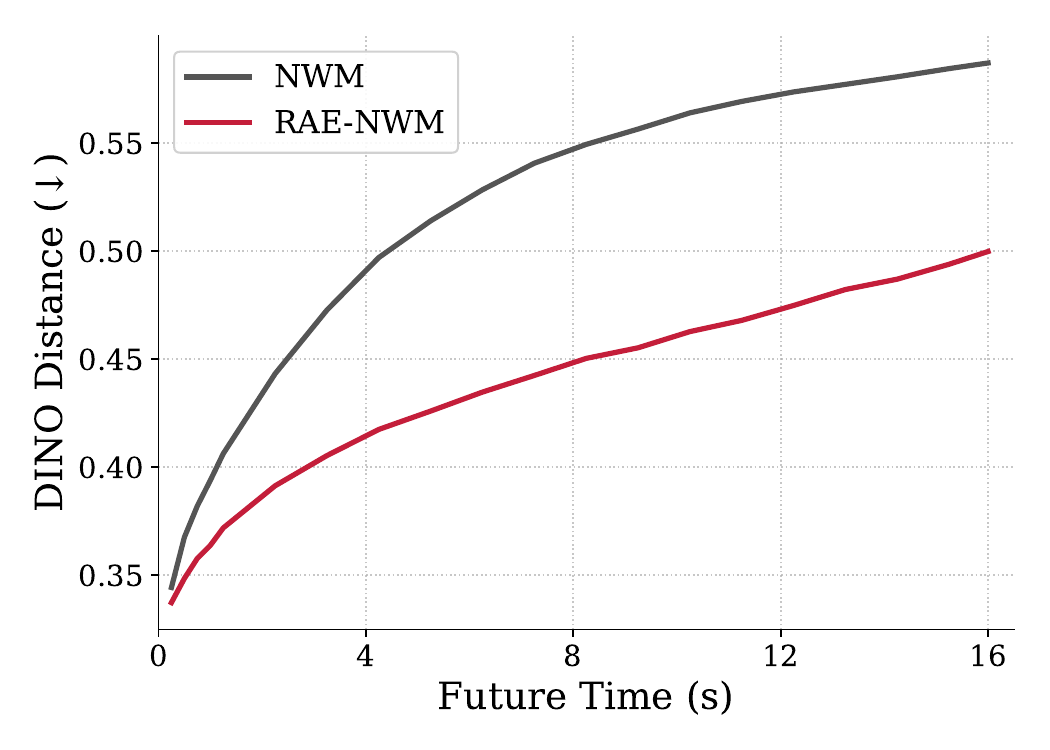}
        \caption{DINO Distance}
        \label{fig:recon_dino}
    \end{subfigure}
    \caption{\textbf{Quantitative performance on the RECON dataset.} Supplementary evaluation of DreamSim and DINO Distance over a 16-second sequential rollout.}
    \label{fig:recon_supp_metrics}
\end{figure}

For the unstructured off-road RECON dataset, we additionally report DreamSim and DINO Distance over the 16-second horizon in Figure~\ref{fig:recon_supp_metrics}, complementing the LPIPS and FID analysis. The results further confirm that RAE-NWM reduces rapid structural degradation compared to the baseline Navigation World Model, maintaining consistent semantic fidelity even in environments with high-frequency stochastic textures.

\begin{figure*}[tbp]
    \centering
    \begin{subfigure}[b]{0.24\textwidth}
        \centering
        \includegraphics[width=\textwidth]{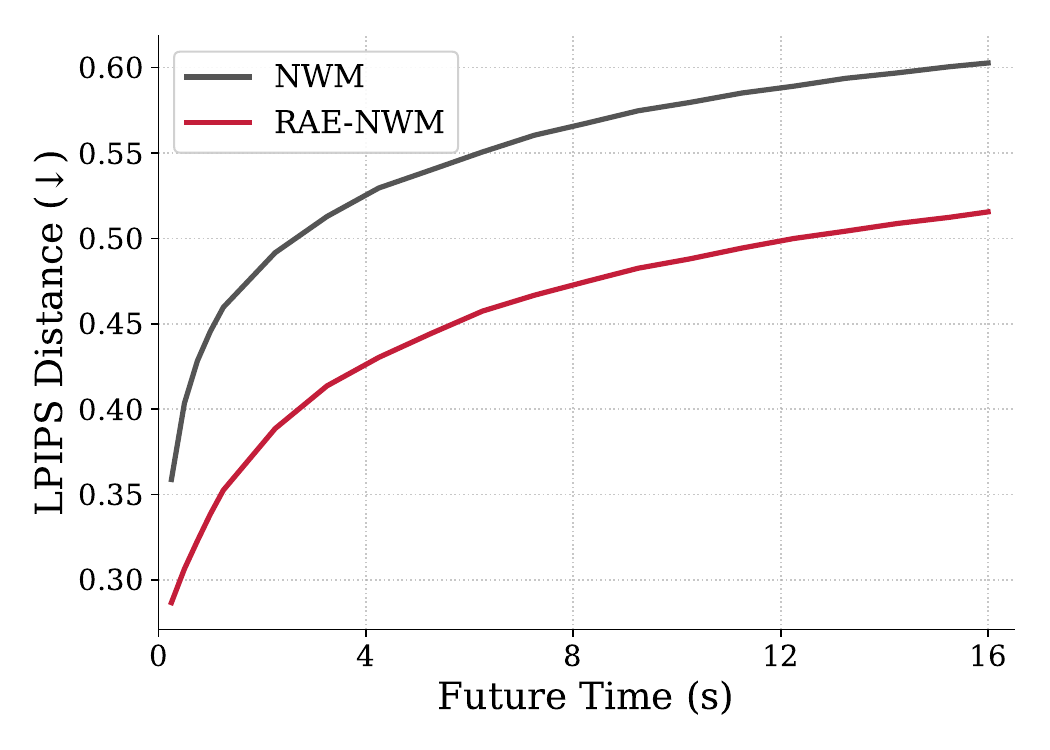}
        \caption{LPIPS}
    \end{subfigure}
    \hfill
    \begin{subfigure}[b]{0.24\textwidth}
        \centering
        \includegraphics[width=\textwidth]{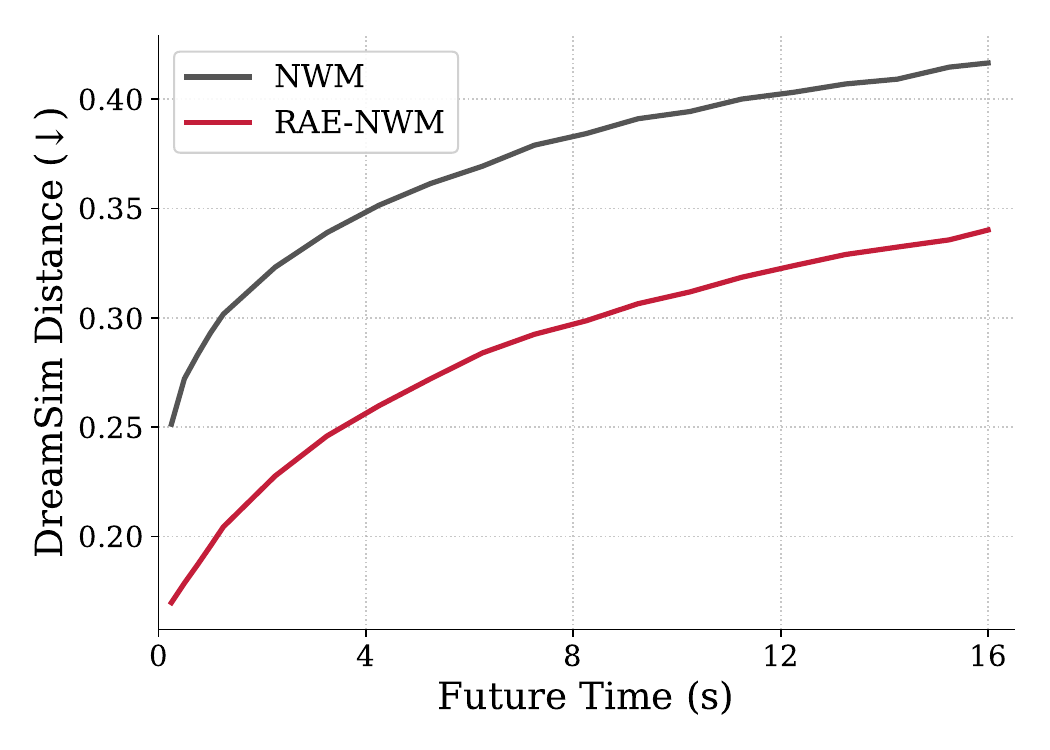}
        \caption{DreamSim}
    \end{subfigure}
    \hfill
    \begin{subfigure}[b]{0.24\textwidth}
        \centering
        \includegraphics[width=\textwidth]{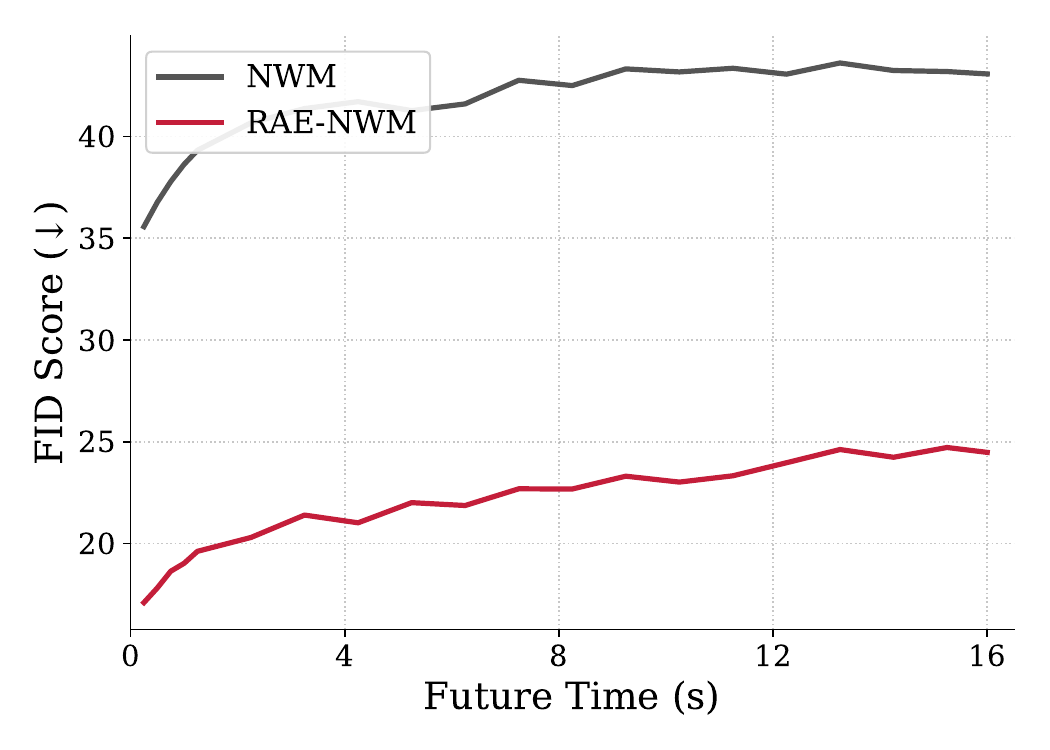}
        \caption{FID}
    \end{subfigure}
    \hfill
    \begin{subfigure}[b]{0.24\textwidth}
        \centering
        \includegraphics[width=\textwidth]{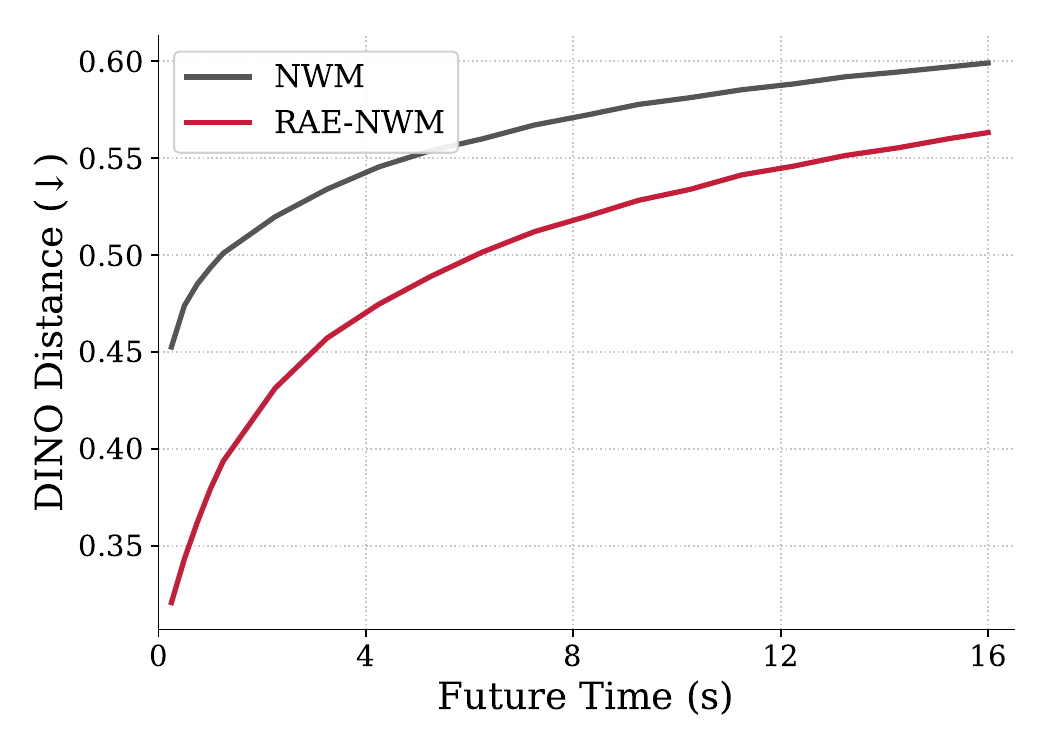}
        \caption{DINO Distance}
    \end{subfigure}
    \caption{\textbf{Quantitative performance on the SCAND dataset.} Performance comparison between RAE-NWM and the baseline over a 16-second sequential rollout.}
    \label{fig:scand_supp_metrics}
\end{figure*}

For the SCAND dataset, which features diverse social navigation demonstrations, we report all four quantitative metrics in Figure~\ref{fig:scand_supp_metrics}. Across perceptual and structural metrics, RAE-NWM exhibits better temporal stability and lower error accumulation over extended horizons compared to the baseline. For this supplementary evaluation, DINO Distance is computed by re-encoding the decoded observations with the frozen DINOv2 encoder, ensuring a consistent image-level comparison across methods.

\subsection{Failure Case Analysis}

\begin{figure}[tbp]
    \centering
    \includegraphics[width=0.6\textwidth]{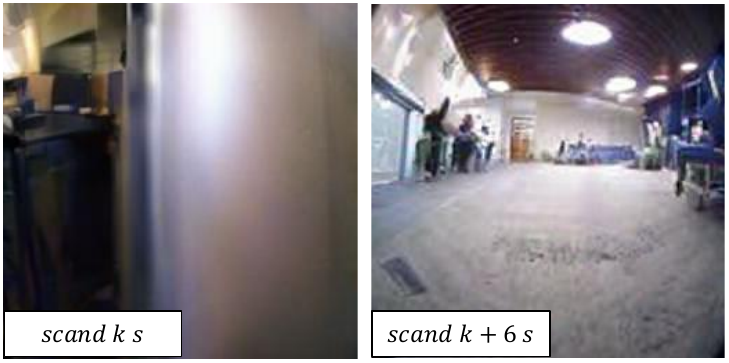}
    \caption{Failure case analysis on the SCAND dataset.}
    \label{fig:failure_case}
\end{figure}

We present a representative failure case on SCAND in Figure~\ref{fig:failure_case}. After collision-induced context ambiguity, RAE-NWM hallucinates a plausible but domain-inconsistent indoor layout. This suggests that dynamic multi-agent scenes and abrupt contact events remain challenging for long-horizon representation-space rollouts.

\subsection{Qualitative Results of Rollout}

We provide qualitative rollout visualizations across four datasets: SACSoN, SCAND, RECON, and Matterport3D. In RECON, fine-grained vegetation details are less faithfully preserved. In SCAND, visual information from the initial context can gradually fade due to dynamic multi-person scenes and limited data scale.

\begin{figure}[tbp]
    \centering
    \includegraphics[width=\textwidth]{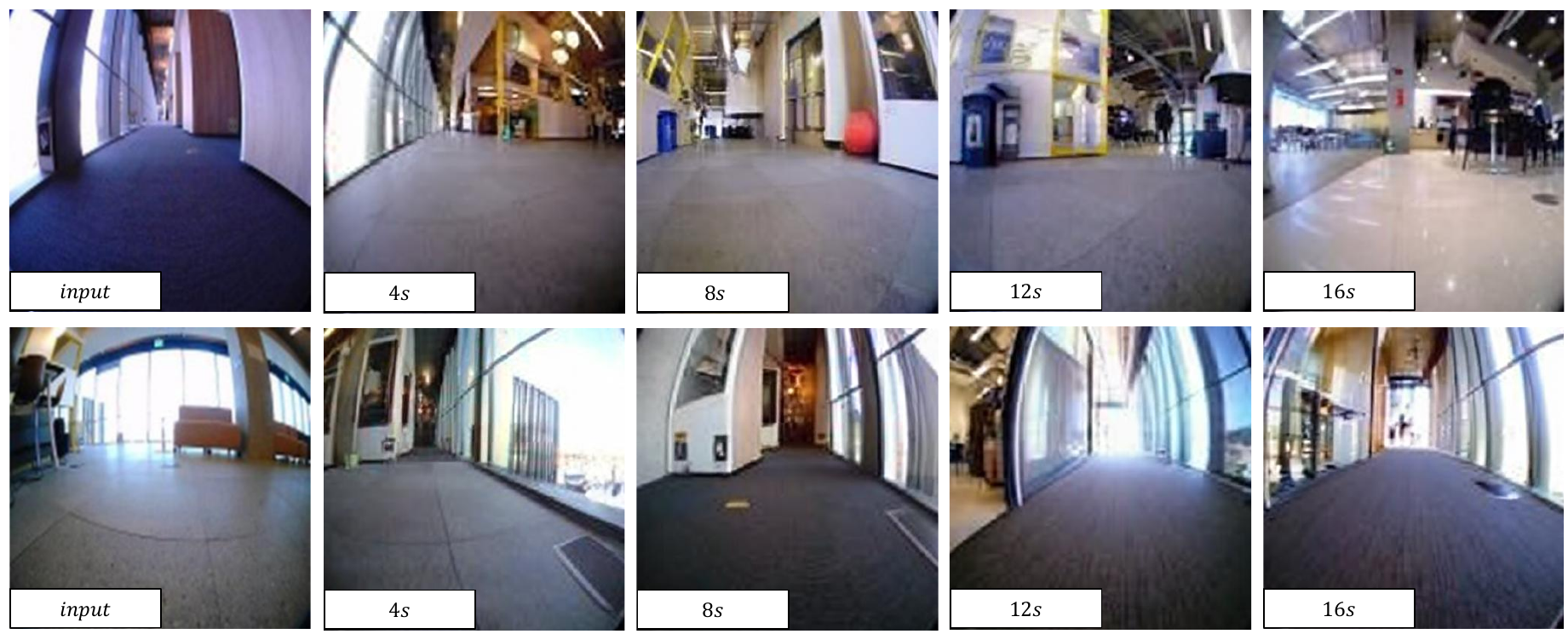}
    \caption{Qualitative rollout results on the SACSoN dataset.}
    \label{fig:sacson_rollout}
\end{figure}

\begin{figure}[tbp]
    \centering
    \includegraphics[width=\textwidth]{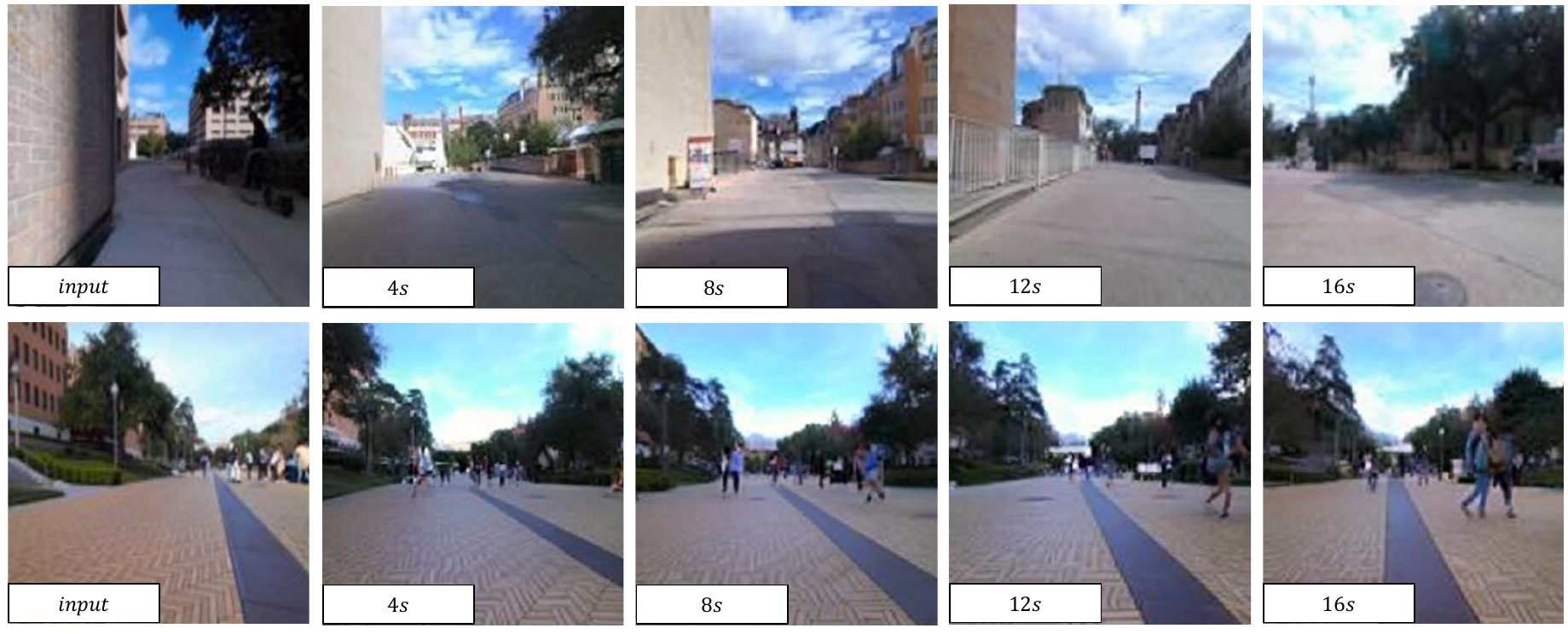}
    \caption{Qualitative rollout results on the SCAND dataset.}
    \label{fig:scand_rollout}
\end{figure}

\begin{figure}[tbp]
    \centering
    \includegraphics[width=\textwidth]{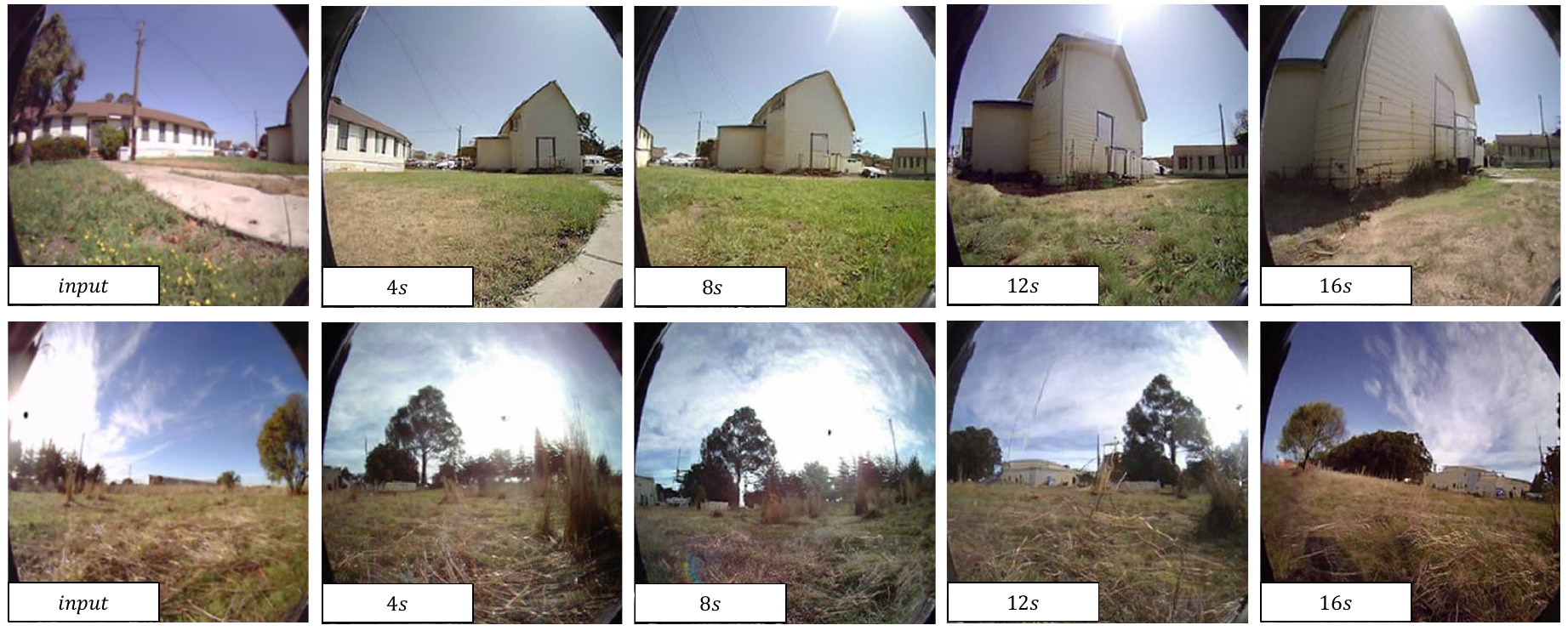}
    \caption{Qualitative rollout results on the RECON dataset.}
    \label{fig:recon_rollout}
\end{figure}

\clearpage
\noindent
\begin{minipage}{\textwidth}
    \centering
    \includegraphics[width=\textwidth]{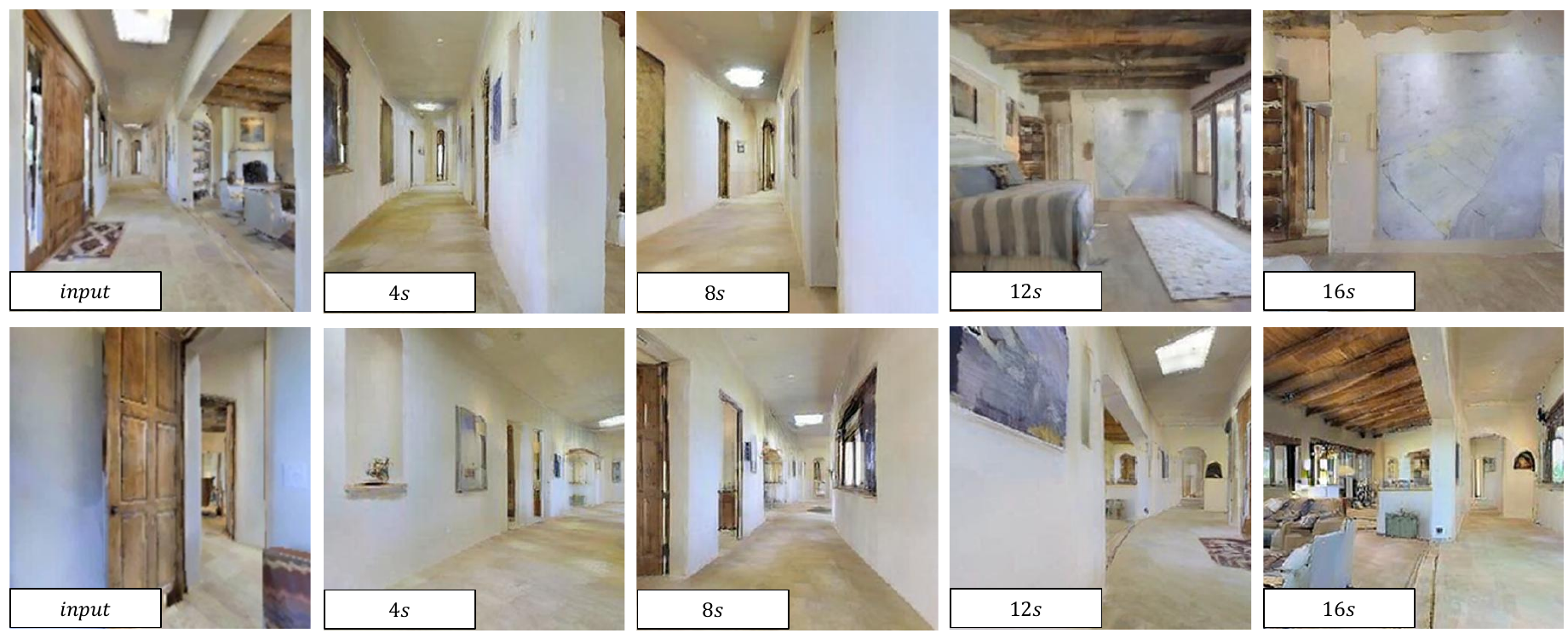}
    \captionof{figure}{Qualitative rollout results on the Matterport3D dataset.}
    \label{fig:mp3d_rollout}
\end{minipage}

\end{document}